\documentclass[journal]{IEEEtran}
\newcommand{\ignore}[1]{}
\usepackage[pass]{geometry}
\usepackage{fancyhdr}
\usepackage[normalem]{ulem}
\usepackage[hyphens]{url}
\usepackage{hyperref}
\usepackage{color}
\usepackage{soul}
\usepackage[sort,compress]{cite}
\usepackage{flushend}
\usepackage{algorithmic}
\usepackage{graphicx}
\usepackage{multirow}
\usepackage{makecell}
\usepackage{amsmath}
\begin{document}
%
\title{Non-Structured DNN Weight Pruning -- Is It Beneficial in Any Platform?}
%
%
%

\author{Xiaolong Ma\text{$^\dagger$},
        Sheng Lin\text{$^\dagger$},
        Shaokai Ye,
        Zhezhi He,
        Linfeng Zhang,
        Geng Yuan,
        Sia Huat Tan,
        Zhengang Li,
        Deliang Fan,
        Xuehai Qian,
        Xue Lin,
        Kaisheng Ma,
        and~Yanzhi Wang
\thanks{$^\dagger$These authors contributed equally.}
}

%
%

\markboth{Journal of \LaTeX\ Class Files,~Vol.~14, No.~8, August~2015}%
{Shell \MakeLowercase{\textit{et al.}}: Bare Demo of IEEEtran.cls for IEEE Journals}
%



\maketitle

\begin{abstract}
Large deep neural network (DNN) models pose
the key challenge to energy efficiency due to
the significantly higher energy consumption of off-chip DRAM accesses than arithmetic or SRAM operations.
It motivates the intensive research on 
model compression with two main approaches. 
Weight pruning leverages the redundancy 
in the number of weights and can be performed 
in a non-structured, which has higher flexibility
and pruning rate but incurs index accesses due to irregular weights, or structured manner, which 
preserves the full matrix structure with lower
pruning rate. 
Weight quantization 
leverages the redundancy in the number 
of bits in weights. 
Compared to pruning, quantization is much more
hardware-friendly, and has become
a ``must-do'' step for FPGA and ASIC implementations.
Thus, any evaluation of the effectiveness 
of pruning should be on top of quantization. 
The {\em key open question} is, 
with quantization, what kind
of pruning (non-structured vs. structured) 
is most beneficial?
This question is fundamental because the answer will
determine the design aspects that we should 
really focus on to avoid diminishing return
of certain optimizations. 



This paper provides a definitive answer to the question for the first time. 
First, we build ADMM-NN-S by extending and enhancing ADMM-NN, a recently 
proposed joint weight pruning and quantization framework, with the algorithmic supports for 
structured pruning, dynamic ADMM regulation, 
and masked mapping and retraining.
Second, we develop a methodology for fair
and fundamental comparison of non-structured
and structured pruning in terms of 
both storage and computation efficiency.
Our results show that ADMM-NN-S
consistently outperforms the prior art:
(i) it achieves 348$\times$, 36$\times$, and 8$\times$ overall weight pruning on LeNet-5, AlexNet, and ResNet-50, respectively, with (almost) zero accuracy loss; (ii) we demonstrate the first fully binarized (for all layers) DNNs can be lossless in accuracy in many cases.
These results provide a strong baseline 
and credibility of our study. 
Based on the proposed comparison framework,
with the same accuracy and quantization,
the results show that non-structured pruning is not 
competitive in terms of both storage 
and computation efficiency.
Thus, we conclude that non-structured pruning
is considered harmful. We urge the community
not to continue the DNN inference acceleration
for non-structured sparsity. 

\end{abstract}


%
\IEEEpeerreviewmaketitle

\section{Introduction}
%
%
%
%
\IEEEPARstart{D}{eep} neural networks (DNNs) with very large
model sizes are the key enabler
for the recent success of deep learning. 
However, large models incur excessive DRAM 
accesses which consume significant more 
energy than arithmetic or SRAM operations. 
Thus, \emph{model compression} of DNNs 
became an active and intensively studied 
research topic.
These techniques, which are applied during the training phase of the DNNs, exploit the 
redundancy in weights.
The aim is to simultaneously reduce the model size (thus, the storage requirement) and accelerate the computation for inference, --- all to be achieved with minor classification accuracy loss. These techniques are 
of particular interests to the hardware acceleration of DNN inference engine
\cite{li2018network,sharma2016high,mao2018lergan,hegde2018morph,chi2016prime,han2016eie,albericio2016cnvlutin,tu2018rana,eckert2018neural,buckler2018eva2,yazdanbakhsh2018ganax,hegde2018ucnn,sharma2018bit,zhang2018pm3,song2019hypar,wang2019bit,liu2015pudiannao,gao2017tetris,ren2017sc,kwon2018maeri,cai2018vibnn,ji2018bridge,zhang2015optimizing,suda2016throughput,qiu2016going,zhao2017accelerating,zhang2017improving,zhang2017frequency,ma2017optimizing,aydonat2017opencl,umuroglu2017finn,gao2018deltarnn,shen2018towards,zeng2018framework,nurvitadhi2018package,chen2018fpga,du2018software,liu2018low,yang2019synetgy,shenaccelerating,jing2019deep,you2019reconfigurable,wei2019overcoming,zhang2019unleashing,zeng2019fine,nakahara2019fpga,lu2019speedy,tang2019ftconv,guo2019compressed,wu2019compute,vogel2019efficient,chetlur2014cudnn,chen2014diannao,judd2016stripes,chen2014dadiannao,venkataramani2017scaledeep,reagen2016minerva,du2015shidiannao,song2018situ,mahajan2016tabla,chen2017eyeriss,moons201714,desoli201714,whatmough201714,sim201614,bang201714,zhang2016caffeine,zhang2016energy,company1,company2}. 
Two important model compression techniques 
are weight pruning and weight quantization.

\begin{figure}
    \centering
    \includegraphics[width=0.35 \textwidth]{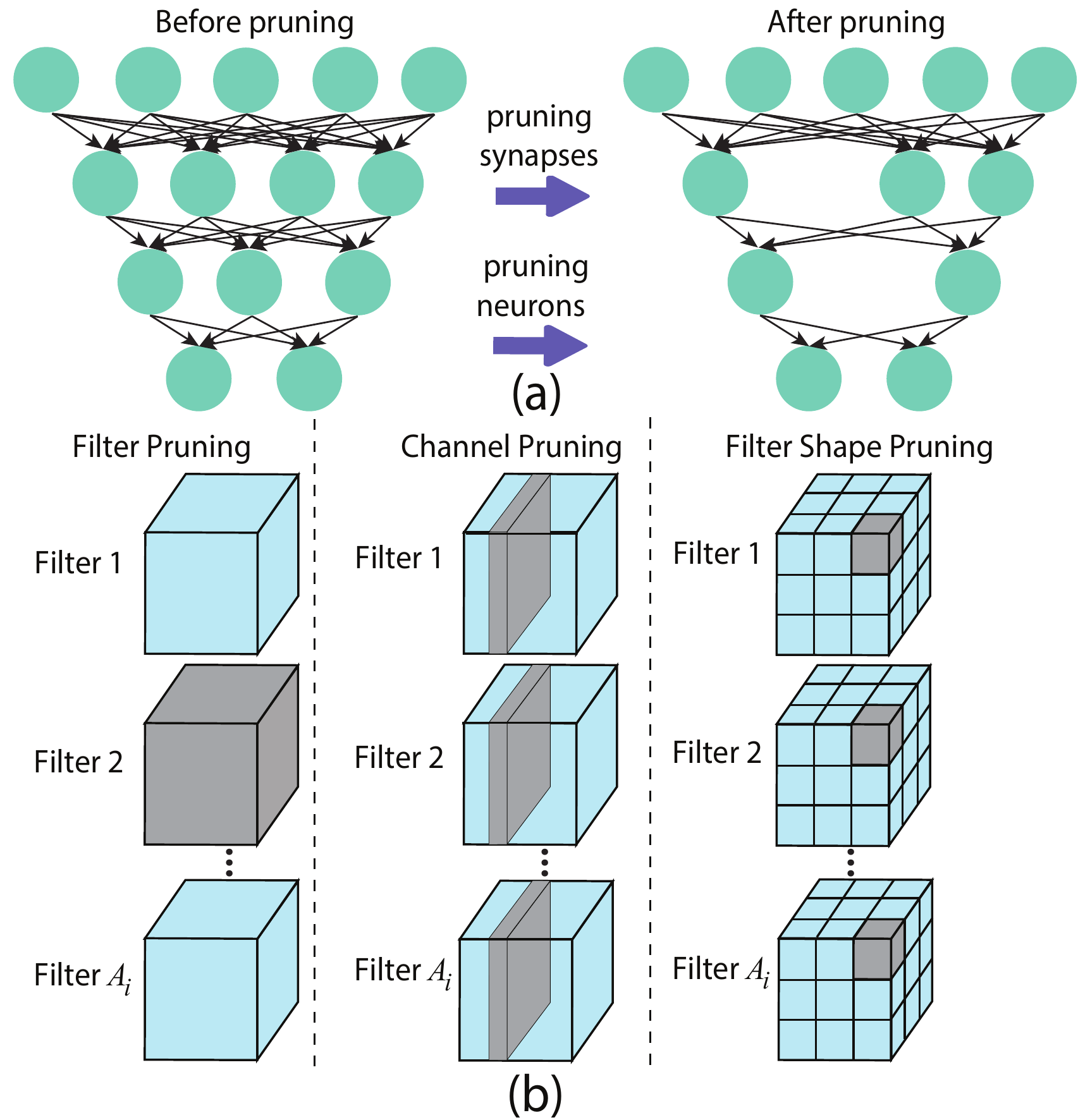}
    \vspace{-2mm}
    \caption{(a) Non-structured weight pruning (arbitrary weight can be pruned) and (b) three types of structured pruning.}
    \label{fig:structuredpruning}
    \vspace{-3mm}
\end{figure}

{\em Weight pruning} leverages the redundancy
in the number of weights. 
One early work~\cite{han2015learning}
used heuristic and iterative weight pruning 
to achieve weight parameter reduction
with negligible accuracy loss. 
It has been extended in \cite{dai2017nest,yang2016designing,guo2016dynamic,dong2017learning} with more sophisticated heuristics. On the downside, such {\em non-structured} methods lead to
{\em irregular, sparse weight matrices} (as 
shown in Figure~\ref{fig:structuredpruning} (a), arbitrary weight can be pruned), 
which rely on 
indices to be stored in a compressed format. 
As a result, they are less compatible with 
the data parallel execution model in GPUs and multicore CPUs.
This drawback is confirmed by the throughput degradation reported in recent works~\cite{wen2016learning,yu2017scalpel}. To overcome the 
limitation of non-structured pruning, recent works~\cite{wen2016learning,he2017channel} proposed the idea of incorporating \emph{regularity or ``structures'' in weight pruning}, such as filter pruning, channel pruning, and filter shape pruning,
shown in Figure~\ref{fig:structuredpruning} (b).
The structured approaches 
maintain a full matrix with reduced dimensions, and {\em indices are no longer needed}. As a result, 
it leads to much higher speedups in GPUs. 

\emph{Weight quantization} is an orthogonal compression technique that leverages the 
redundancy in the number of bits of 
weight representation \cite{leng2017extremely,park2017weighted,zhou2017incremental,lin2016fixed,wu2016quantized,rastegari2016xnor,hubara2016binarized,courbariaux2015binaryconnect}. 
Compared to weight pruning, 
weight quantization is inherently more hardware-friendly, 
since both storage and computation of DNNs will be reduced proportionally to the weight precision
without additional overhead due to indices.
Moreover, multiplication operations 
may be eliminated with binary, ternary, or power-of-2 weight quantizations \cite{rastegari2016xnor,hubara2016binarized,courbariaux2015binaryconnect}. 
Thanks to these advantages, weight quantization 
has been a ``must-do'' step for DNN inference engines.
Besides FPGA and ASIC, it is also well supported in GPU, CPU, and mobile devices, e.g., \cite{TensorFlow-Lite,paszke2017pytorch}.

\begin{figure}
    \centering
    \includegraphics[width=0.45 \textwidth]{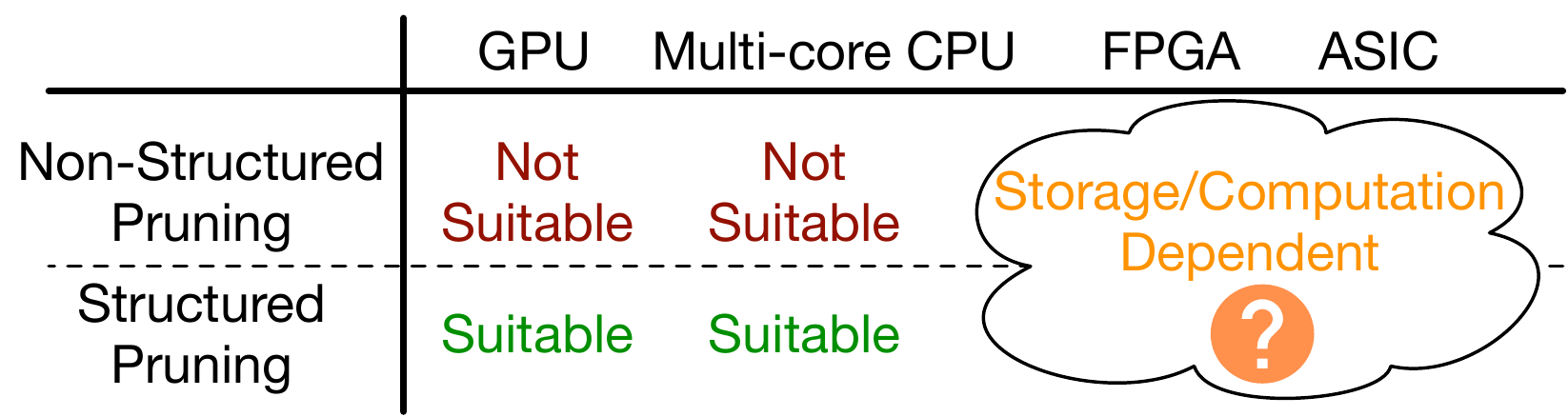}
    \vspace{-2mm}
    \caption{Is non-structured pruning beneficial at all?}
    \label{fig:questionofsparsity}
    \vspace{-3mm}
\end{figure}

Given the pros and cons of non-structured/structured
weight pruning and weight quantization, 
they need to be investigated jointly to
fully understand the interactions between them.
In particular, since weight quantization is a must-do step, especially for FPGA and ASIC, i.e., weight pruning will not be performed alone. 
The {\bf key open question} is, {\em with quantization,
what kind of pruning (non-structured vs. structured) is most beneficial}?
The answer to the question is far from obvious.
Using LeNet-5 (for MNIST data set) as an example, we achieve an unprecedented 348$\times$ (non-structured) weight reduction with 3-bit quantization, maintaining 99$\%+$ accuracy. However, each index needs to be at least 9-bit on account of 348$\times$ weight pruning. 
This makes index storage larger than that of weights (in addition, 
indices cannot be further quantized).
In this example, 
non-structured weight pruning results in larger actual storage than structured pruning.
Thus, we can see the importance of answering
such question: it will determine
the design aspects that 
we should really focus on to avoid diminishing 
return of certain optimizations. As shown in Figure \ref{fig:questionofsparsity}, we need 
answers for all platforms.

Two 
recent works ADMM-NN~\cite{admm2019asplos} and \cite{leng2017extremely}, that perform systematic joint weight pruning and quantization, are
in the best position to perform this study. 
Using advanced variable-splitting optimization 
method ADMM (Alternating Direction Methods of Multipliers)~\cite{ouyang2013stochastic,suzuki2013dual,boyd2011distributed}, 
state-of-the-art results are achieved (e.g., 21$\times$ weight reduction \cite{zhang2018systematic} in AlexNet), ---  
outperforming heuristic counterparts.
Unfortunately, the current framework is {\em insufficient} to 
perform such study. 
First, ADMM-NN lacks the algorithmic mechanisms
to enforce structured weight pruning, and 
guarantee the solution feasibility.
Second, we lack the methodology to fairly and fundamentally
compare non-structured and structured 
pruning in an ``apple-to-apple'' manner.
This paper is the {\em first} study 
to provide the answer to the open question 
with two key contributions. 


The {\bf first contribution} of the paper is the
development of ADMM-NN-S by
extending and enhancing of ADMM-NN~\cite{admm2019asplos}.
It is extended with the algorithmic supports
for structured pruning. We achieve this by 
adjusting the constraints in each layer to 
express the structured requirements. 
For example, for filter pruning, the constraint
for a layer can be specified as
{\em number of non-zero
filters} is less than or equal to a threshold. 
Moreover, we develop a systematic framework
of dynamic ADMM regulation, masked mapping 
and retraining to guarantee solution 
feasibility (satisfying all constraints)
and provide high solution quality
(ensuring pruning and quantization rate 
under the same accuracy).

The {\bf second contribution} is the 
methodology for the fair and fundamental 
comparison of non-structured and structured 
weight pruning with quantization in place. 
We focus on two metrics with the {\em same accuracy}:
1) total storage (weight+indices), which is computed based on both absolute and relative indices;
2) computation efficiency, which is captured by 
a new metrics called pruning-to-performance ratio 
(PPR). After pruning, suppose $\alpha\times$ weight reduction results in $\beta\times$ speedup, 
the PPR value is defined as $\alpha/\beta$.
Intuitively, the less the value of PPR,
the higher the computation efficiency, 
--- same speedup can be achieved by smaller 
pruning rate. 
For structured pruning, PPR value is 
approximately 1 due to the absence of indices.
For non-structured pruning, recent accelerators
based on non-structured 
sparsity~\cite{yuan2018sticker,ren2018admm,zhang2016cambricon,parashar2017scnn} show that 
PPR values are larger than 2.7.
We can fairly compare non-structured and structured
pruning by conservatively comparing PPR:
non-structured pruning is more beneficial 
if it can achieve 2.7$\times$
or higher pruning rate than structured pruning.
No prior work has conducted such study and the 
answer to the above comparison is {\em unknown}. 

The fairness of the proposed methodology
is ensured due to three reasons: 
1) it is performed by our new ADMM-NN-S
framework that 
significantly {\em outperforms} prior arts (in both non-structured and structured pruning); 
2) the comparison of storage and computation is \emph{hardware implementation-agnostic};
3) the comparison is performed at \emph{the same rate of accuracy}. We also strengthen weight quantization after non-structured pruning by selectively leveraging state-of-art ternary quantization solution \cite{he2018simultaneously}.

Based on the proposed ideas, we perform
extensive and representative 
testing of our comparison framework 
with AlexNet, VGGNet, ResNet-18/50, MobileNet, and LeNet-5 models based on ImageNet, CIFAR-10, and MNIST data sets. Due to space limitation, we focus on convolutional (CONV) layers, which are the most computationally intensive layers in DNNs and are becoming the major storage as well as in state-of-art ResNet and MobileNet models. We do observe similar (and more 
significant) effect on fully-connected (FC) layers and on RNNs.
We highlight our results and findings.

First, ADMM-NN-S framework guarantees solution feasibility while providing high solution quality.
Our results consistently and significantly outperform prior art.
This is the key to ensure the credibility
of our conclusion.
Specifically, we 
1) achieve {\em unprecedented} 348$\times$, 36$\times$, and 8$\times$ overall weight pruning on LeNet-5, AlexNet, and ResNet-50 models, respectively, with (almost) zero accuracy loss; 
2) derive the first lossless, fully binarized (for all layers) LeNet-5 for MNIST and VGG-16 for CIFAR-10; and 3) derive the first fully binarized (for all layers) ResNet for ImageNet with reasonable accuracy loss.

Second, comparing non-structured and structured
pruning, we find that the storage overhead of indices for non-structured pruning 
is always more than its additional weight storage reduction,
thus the amount of total storage for non-structured pruning is actually larger.
In term of computation efficiency, 
we find that the PPR for structured pruning
in all models are less than 2.7$\times$. 
For the first time, our results
show that, despite more flexibility and 
weight pruning rate, {\em non-structured pruning is not competitive
in terms of both storage and computation efficiency}
with quantization and the same accuracy. 
In a few cases, the storage size of non-structured pruning is comparable (or slightly better than) to that of structured pruning, however it is still not a desirable choice considering the additional complexity of hardware design to support non-structured sparsity. 
As a result, we 
reach the conclusion that 
{\bf non-structured weight pruning is 
considered harmful}, and we 
recommend not to continue investigating DNN inference engines using non-structured sparsity.
We release codes and all the models of this work at anonymous link: \url{http://bit.ly/2WMQSRi}.

\section{Model Compression Background}

\subsection{Weight Pruning}

\begin{figure*}[thb]
    \centering
    \includegraphics[width=0.75 \textwidth]{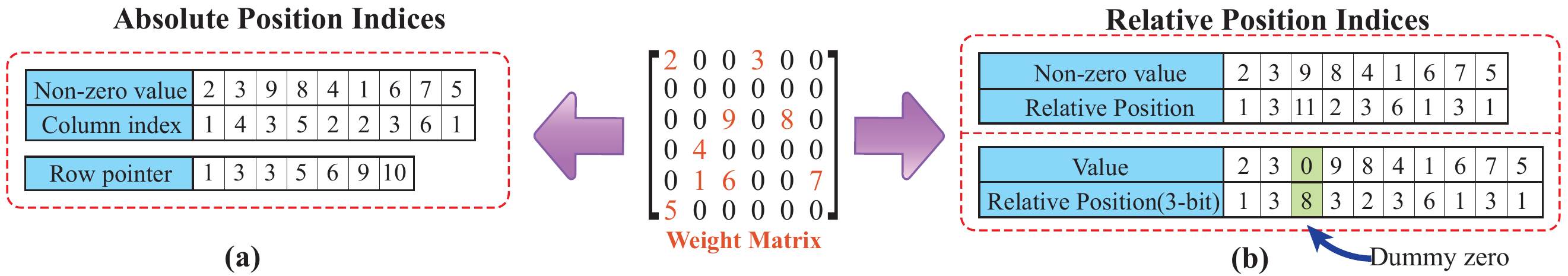}
    \vspace{-1mm}
    \caption{Compressed sparse row (CSR) format with (a) absolute indices and (b) relative indices.}
    \label{fig:CSR}
    \vspace{-3mm}
\end{figure*}

\textbf{\emph{Non-structured weight pruning.}} The early work by Han \emph{et al.} \cite{han2015learning} achieved 9$\times$ reduction in the number of parameters in AlexNet and 13$\times$ in VGG-16. However, most reduction is achieved in FC layers, and the 2.7$\times$ reduction achieved in CONV layers will not lead to an overall acceleration in GPUs \cite{wen2016learning}. 
Extensions of iterative weight pruning, such as \cite{guo2016dynamic} (dynamic network surgery), \cite{dai2017nest} (NeST) and \cite{mao2017exploring}, use more delicate algorithms such as selective weight growing and pruning. But the weight pruning rates on CONV layers are still limited, e.g., 3.1$\times$ in \cite{guo2016dynamic}, 3.23$\times$ in \cite{dai2017nest}, and 4.16$\times$ in \cite{mao2017exploring} for AlexNet with no accuracy degradation. This level of non-structured weight pruning cannot guarantee sufficient speedups in GPUs. In fact, based on the enhanced ADMM-NN
framework, we can achieve 11.2$\times$ non-structured weight pruning in CONV layers with almost no accuracy degradation. Ironically, it even results in {\em 20\% speed degradation} on an NVIDIA 1080Ti GPU.

\textbf{\emph{Structured weight pruning.}}
To overcome the limitation in non-structured, irregular weight pruning, SSL \cite{wen2016learning} proposes to learn structured sparsity at the levels of filters, channels, filter shapes, layer depth, etc. This work is among the firsts that 
reported the actually measured GPU accelerations. This is because CONV layers after structured pruning will transform to a full matrix multiplication with reduced matrix size. 
However, the weight pruning rate is limited in the prior work on structured pruning. 
The average weight pruning rate on CONV layers of AlexNet is only 1.4$\times$ without accuracy loss. More recently, \cite{he2017channel} achieved 2$\times$ channel pruning with 1\% accuracy degradation on VGGNet.
More importantly, the structured 
pruning has never been evaluated 
with weight quantization.

\subsection{Weight Quantization}
\textbf{\emph{Weight quantization.}} This method takes advantages of the inherent redundancy in the number of bits for weight representation. 
Many of the prior works \cite{leng2017extremely,park2017weighted,zhou2017incremental,lin2016fixed,wu2016quantized,rastegari2016xnor,hubara2016binarized,courbariaux2015binaryconnect} focused on quantization of weights to binary values, ternary values, or powers of 2 to facilitate hardware implementation, with acceptable accuracy loss. 
The state-of-the-art techniques \cite{courbariaux2015binaryconnect,leng2017extremely} adopt an iterative quantization and retraining framework, with some degree of randomness incorporated into the quantization step. 
This method results in less than 3\% accuracy loss on AlexNet for binary weight quantization \cite{leng2017extremely}. 

Compared to weight pruning, weight quantization is the major DNN model compression technique utilized in industry, due to its ``hardware-friendliness''
and the proportional reduction of computation and storage. 
Thus, weight quantization has been
a must-do step in FPGA and ASIC designs of DNN inference engines. Also, it is well supported in GPUs and mobile devices, e.g., PyTorch \cite{paszke2017pytorch} in NVIDIA GPU and TensorFlow Lite \cite{TensorFlow-Lite} for mobile devices.

\subsection{ADMM for Weight Pruning/Quantization}

Recent work \cite{admm2019asplos,leng2017extremely} have incorporated ADMM for DNN weight pruning and weight quantization, respectively. ADMM is a powerful tool for optimization, by decomposing an original problem into two subproblems that can be solved separately and efficiently. For example, considering optimization problem $\min_{\bf{x}} f({\bf{x}})+g({\bf{x}}).$ In ADMM, this problem is decomposed into two subproblems on $\bf{x}$ and $\bf{z}$ (auxiliary variable), which will be solved iteratively until convergence. The first subproblem derives $\bf{x}$ given $\bf{z}$: $\min_{\bf{x}} f({\bf{x}})+q_1(\bf{x}|\bf{z})$. The second subproblem derives $\bf{z}$ given $\bf{x}$: $\min_{\bf{z}} g({\bf{z}})+q_2(\bf{z}|\bf{x})$. Both $q_1$ and $q_2$ are quadratic functions. 

ADMM is conventionally utilized to accelerate the convergence of convex optimization problems and enable distributed optimization, in which the optimality and fast convergence rate has been proven \cite{ouyang2013stochastic,boyd2011distributed}. As a special property, ADMM can effectively deal with a subset of combinatorial constraints and yields optimal (or at least high quality) solutions \cite{hong2016convergence,liu2018zeroth}. Luckily, the associated constraints in the DNN weight pruning and quantization belong to this subset of combinatorial constraints, 
making ADMM applicable to DNN mode compression.
However, due to the non-convex nature of the objective function for DNN training, there is still a lack of guarantee in the prior work \cite{admm2019asplos,leng2017extremely} on \emph{solution feasibility} and \emph{solution quality}.
Moreover, ~\cite{admm2019asplos} only supports non-structured pruning.

\section{Non-Structured vs. Structured Weight Pruning}

\subsection{Non-Structured Pruning: Indexing Overhead}

Indices are used to represent weight matrices in the sparse format, thereby achieving storage reduction in non-structured weight pruning. A representative sparse representation format is the \emph{compressed sparse row} (CSR) format, which was also utilized in prior work \cite{han2015learning,han2016eie}. As shown in \textcolor{black}{Figure \ref{fig:CSR} (a)}, it represents a matrix by three arrays, which respectively contains nonzero (weight) values, column indices and the extents of rows. This representation requires $2n + r + 1$ numbers, where $n$ is the number of nonzero values and $r$ is the number of rows.

We call the above representation as \emph{CSR with absolute indices}.
Instead of storing the absolute position, we can compute the index difference and store the indices with relative position. This representation requires $2n$ numbers, where $n$ is the number of nonzero (weight) values. For further compression, one can restrict the number of bits (3 bits in this example) to represent the relative position and add a dummy zero weight when the relative position exceeds the largest value (8 for this example) that can be represented, both shown in \textcolor{black}{Figure \ref{fig:CSR} (b)}. These cases are called \emph{CSR with relative indices}. 


Comparing the two options, CSR with relative indices is good for compression \cite{han2015learning}, while CSR with absolute indices leads to better hardware acceleration \cite{yuan2018sticker,zhang2016cambricon,parashar2017scnn}. In this work, we aim to allow the highest freedom for non-structured pruning in storage and computation evaluations, --- we allow CSR with relative indices in storage calculation and CSR with absolute indices for computation estimation for non-structured pruning.

\subsection{Structured Pruning: Three Types}\label{sec:structuredpruning}

\begin{figure}
    \centering
    \includegraphics[width=0.45 \textwidth]{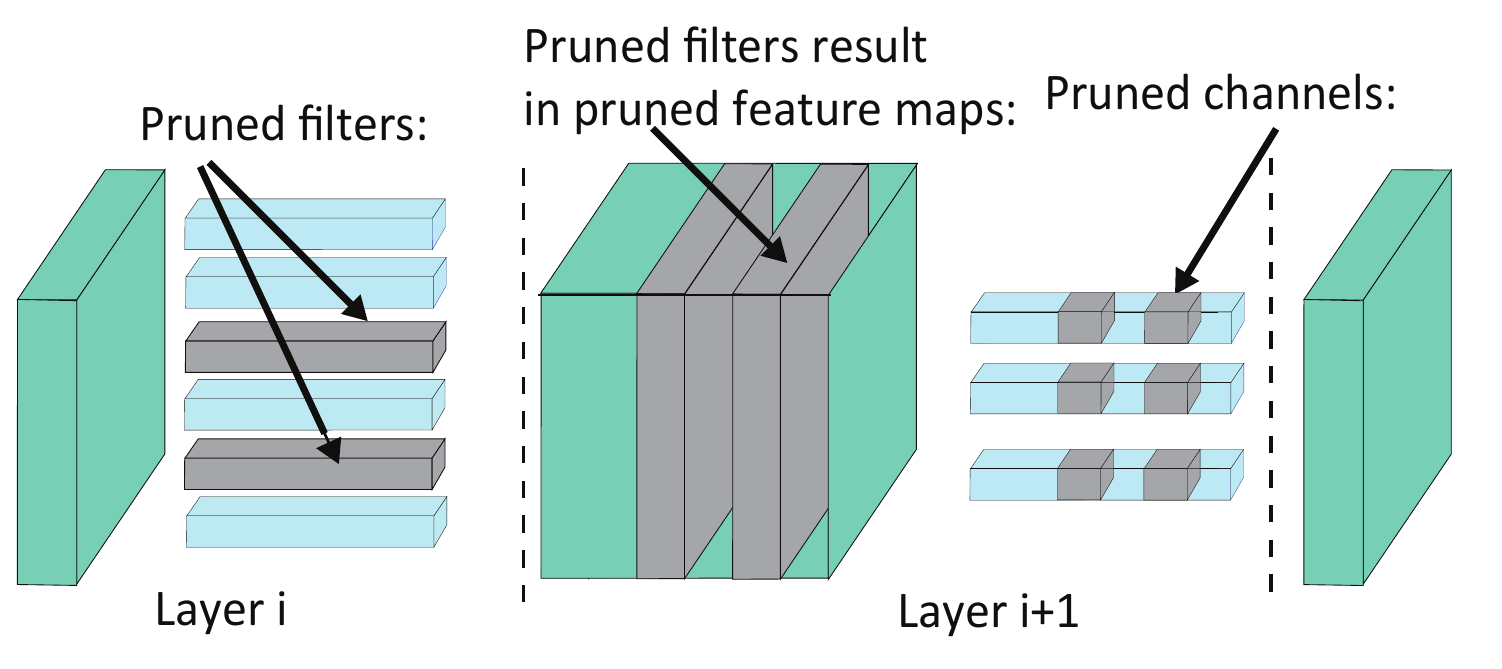}
    \vspace{-4mm}
    \caption{Relation between filter pruning and channel pruning. Pruned filters in layer $i$ results in pruned feature maps and therefore pruned (inactivated) channels in layer $i+1$.}
    \label{fig:filterchannel}
    \vspace{-3mm}
\end{figure}

Wen {\em et al.}~\cite{wen2016learning} introduced 
three types of structured pruning: \emph{filter pruning}, \emph{channel pruning}, and \emph{filter shape pruning}, as shown in \textcolor{black}{Figure \ref{fig:structuredpruning} (b)}. Filter pruning removes whole filter(s); channel pruning removes whole channels; and filter shape pruning removes the weights in the same locations of all filters in one specific layer. Moreover, as shown in \textcolor{black}{Figure \ref{fig:filterchannel}}, filter pruning and channel pruning are correlated. Pruning a filter in layer $i$ is equivalent to pruning the corresponding channel in layer $i+1$, which is generated by this specific filter. As a result, filter pruning (and channel pruning) has a roughly quadratic effect on the weight parameter reduction (and the amount of computations) of the DNNs.

\begin{figure}
    \centering
    \includegraphics[width=0.43 \textwidth]{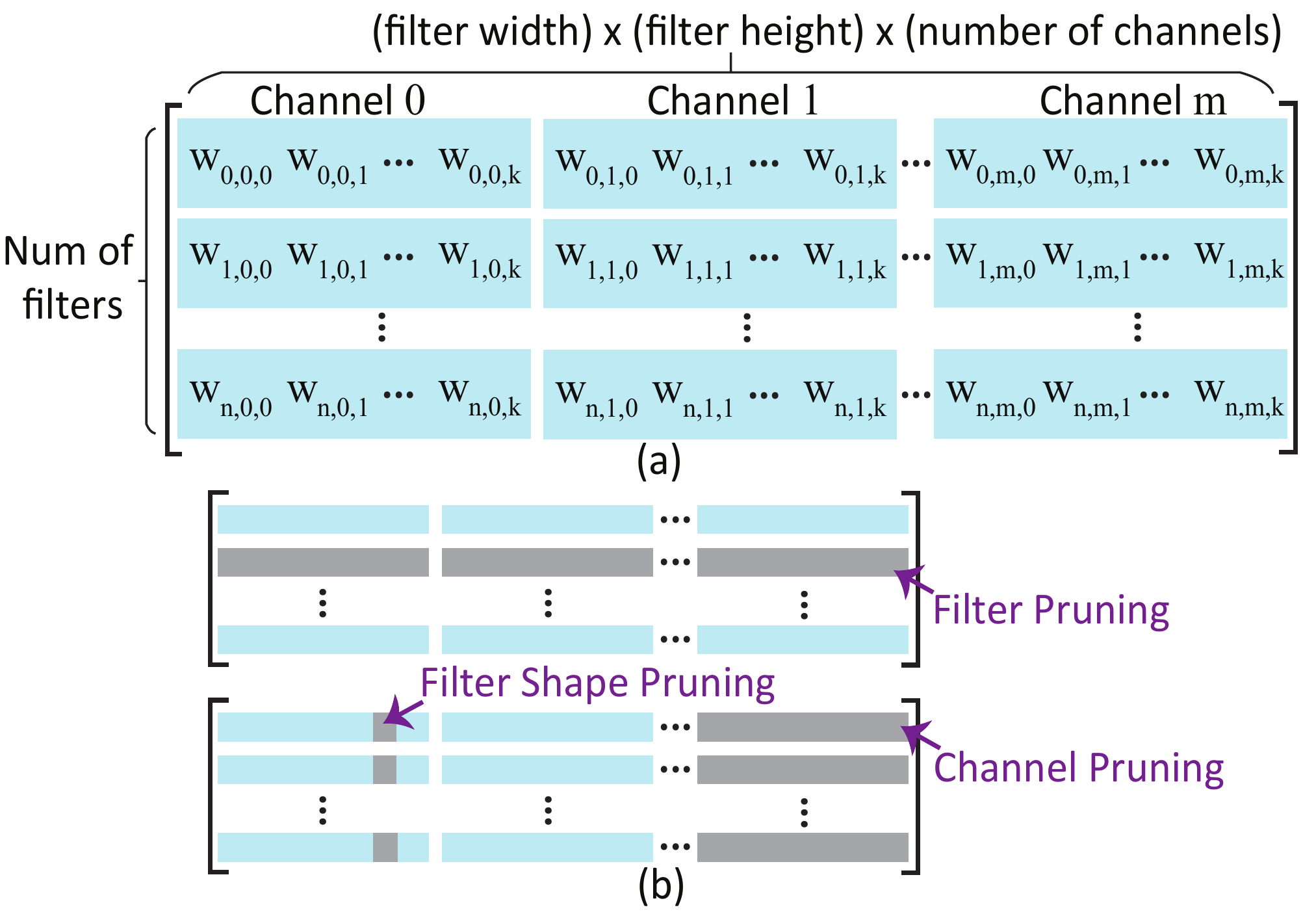}
    \caption{(a) To support GEMM, weight tensor representation of a CONV layer is transformed into weight matrix representation. (b) How different structured weight pruning schemes are implemented on weight matrix representation.}
    \label{fig:GEMM}
\end{figure}

The CONV operations in (one layer of) DNNs are commonly transformed to matrix multiplications by converting weight tensors and feature map tensors to matrices \cite{chetlur2014cudnn}, named \emph{general matrix multiplication} or GEMM, as shown in \textcolor{black}{Figure \ref{fig:GEMM}}.
From \textcolor{black}{Figure \ref{fig:GEMM} (b)}, filter pruning corresponds to reducing one row, and thus is also termed \emph{row pruning}. Filter shape pruning corresponds to reducing one column, and thus is also termed \emph{column pruning}. Channel pruning corresponds to reducing multiple consecutive columns. The three structured pruning techniques, along with their combinations, will reduce the dimensions in GEMM while maintaining a full matrix format.
Thus, indices are not needed.
It is why structured pruning is in general 
more suitable for hardware accelerations.

On one hand, the major advantage of filter/channel pruning has the superlinear effect on storage/computation reduction, i.e., $\alpha\times$ filter pruning on all layers results in over $\alpha\times$ reduction in number of weight parameters. On the other hand, column pruning has a higher degree of flexibility. These techniques can be largely combined in order to achieve the highest rates in reductions of computation and storage, and effective heuristic for the desirable combination is needed.



\section{ADMM-NN-S Framework}

In this section, we build ADMM-NN-S, a unified solution framework of both non-structured and structured weight pruning, as well as weight quantization problems by extending ADMM-NN,
the state-of-the-art ADMM-based framework~\cite{admm2019asplos}. 
The differences between ADMM-NN-S and 
ADMM-NN are:
1) it supports structured pruning;
2) it can guarantee solution feasibility and provide high solution quality; and 
3) we propose effective techniques for enhancing convergence.

\subsection{Enforcing Structured Pruning}

This section discusses the extension of 
ADMM-NN with structured pruning constraints. 
Consider an $N$-layer DNN with both CONV and FC layers. The weights and biases of the $i$-th layer are respectively denoted by ${\bf{W}}_{i}$ and ${\bf{b}}_{i}$, and the loss function associated with the DNN is denoted by $f \big( \{{\bf{W}}_{i}\}_{i=1}^N, \{{\bf{b}}_{i} \}_{i=1}^N \big)$; see \cite{zhang2018systematic}. In our 
discussion, $\{{\bf{W}}_{i}\}_{i=1}^N$ and $\{{\bf{b}}_{i} \}_{i=1}^N$ respectively characterize the collection of weights and biases from layer $1$ to layer $N$. Then DNN weight pruning or weight quantization is formulated as optimization problem:
\begin{equation}
\label{opt0}
\begin{aligned}
& \underset{ \{{\bf{W}}_{i}\},\{{\bf{b}}_{i} \}}{\text{minimize}}
& & f \big( \{{\bf{W}}_{i}\}_{i=1}^N, \{{\bf{b}}_{i} \}_{i=1}^N \big),
\\ & \text{subject to}
& & {\bf{W}}_{i}\in {\mathcal{S}}_{i}, \; i = 1, \ldots, N,
\end{aligned}
\end{equation}

Next we introduce constraint sets ${\mathcal{S}}_{i}$'s corresponding to the non-structured weight pruning, different types of structured pruning, as well as weight quantization. We use CONV layers as illustrative example since CONV layers are the most computationally intensive. The problem formulation can be well applied to FC layers \cite{zhang2018systematic}.

The collection of weights in the $i$-th CONV layer is a four-dimensional tensor, i.e., ${\bf{W}}_{i} \in R^{A_i \times B_i \times C_i \times D_i}$, where $A_i, B_i, C_i$, and $D_i$ are respectively the number of filters, the number of channels in a filter, the height of the filter, and the width of the filter, in layer $i$.
In the following,
if $\bf{X}$ denotes the weight tensor in a specific layer, let $({\bf{X}})_{a,:,:,:}$ denote the $a$-th filter in $\bf{X}$, $({\bf{X}})_{:,b,:,:}$ denote the $b$-th channel, and $({\bf{X}})_{:,b,c,d}$ denote the collection of weights located at position $(:,b,c,d)$ in every filter of $\bf{X}$,
as illustrated in \textcolor{black}{Figure \ref{fig:structuredpruning} (b)}.

\emph{\textbf{Weight pruning}}: For \emph{non-structured weight pruning}, the constraint on the weights in $i$-th layer is
$
{\bf{W}}_{i} \in {\mathcal{S}}_{i} := \{{\bf{X}}\mid$ {number of nonzero elements in} ${\bf{X}}$ {is less than or equal to} $\alpha_{i}  \}.$
For \emph{filter pruning} (row pruning), the constraint in the $i$-th CONV layer becomes ${\bf{W}}_{i}\in {\mathcal{S}}_{i}
:=
\{{\bf{X}}\mid$ {the number of nonzero filters in} ${\bf{X}}$ {is less than or equal to} $\beta_i \}$.
For \emph{channel pruning}, the constraint becomes ${\bf{W}}_{i} \in {\mathcal{S}}_{i} :=  \{{\bf{X}}\mid$ {the number of nonzero channels in} ${\bf{X}}$ {is less than or equal to} $\gamma_{i}  \}.$
Finally, for \emph{filter-shape pruning} (column pruning), the constraint in the $i$-th CONV layer is ${\bf{W}}_{i}\in {\mathcal{S}}_{i}
:=
 \{{\bf{X}}\mid$ {the number of nonzero vectors in} $\{{\bf{X}}_{:,b,c,d}\}_{b,c,d=1}^{B_i,C_i,D_i}$ {is less than or equal to} $\theta_i  \}.$ These $\alpha_i$, $\beta_i$, $\gamma_i$, and $\theta_i$ values are hyperparameters determined in prior, and the determination procedure will be discussed in Section \ref{sec:hyper_deter}.

\emph{\textbf{Weight quantization}}: For weight quantization, elements in ${\bf{W}}_{i}$ assume one of  $q_{i,1},q_{i,2},...,q_{i,M_i}$ values, where $M_i$ denotes the number of these fixed values. The $q_{i,j}$ values are \emph{quantization levels} of weights of layer $i$ in increasing order, and we focus on \emph{equal-distance quantization} (the same distance between adjacent quantization levels) to facilitate hardware implementation.

\subsection{Enhancing Solution Feasibility and High Solution Quality}

In problem (\ref{opt0}), the constraint is combinatorial. 
As a result, this problem cannot be solved directly by stochastic gradient descent methods like original DNN training.
However, the form of the combinatorial constraints on ${\bf{W}}_{i}$ is compatible with ADMM which is recently shown to be an effective method to deal with such clustering-like constraints \cite{hong2016convergence,liu2018zeroth}.

\begin{figure} 
     \centering
     \includegraphics[width=0.6\columnwidth]{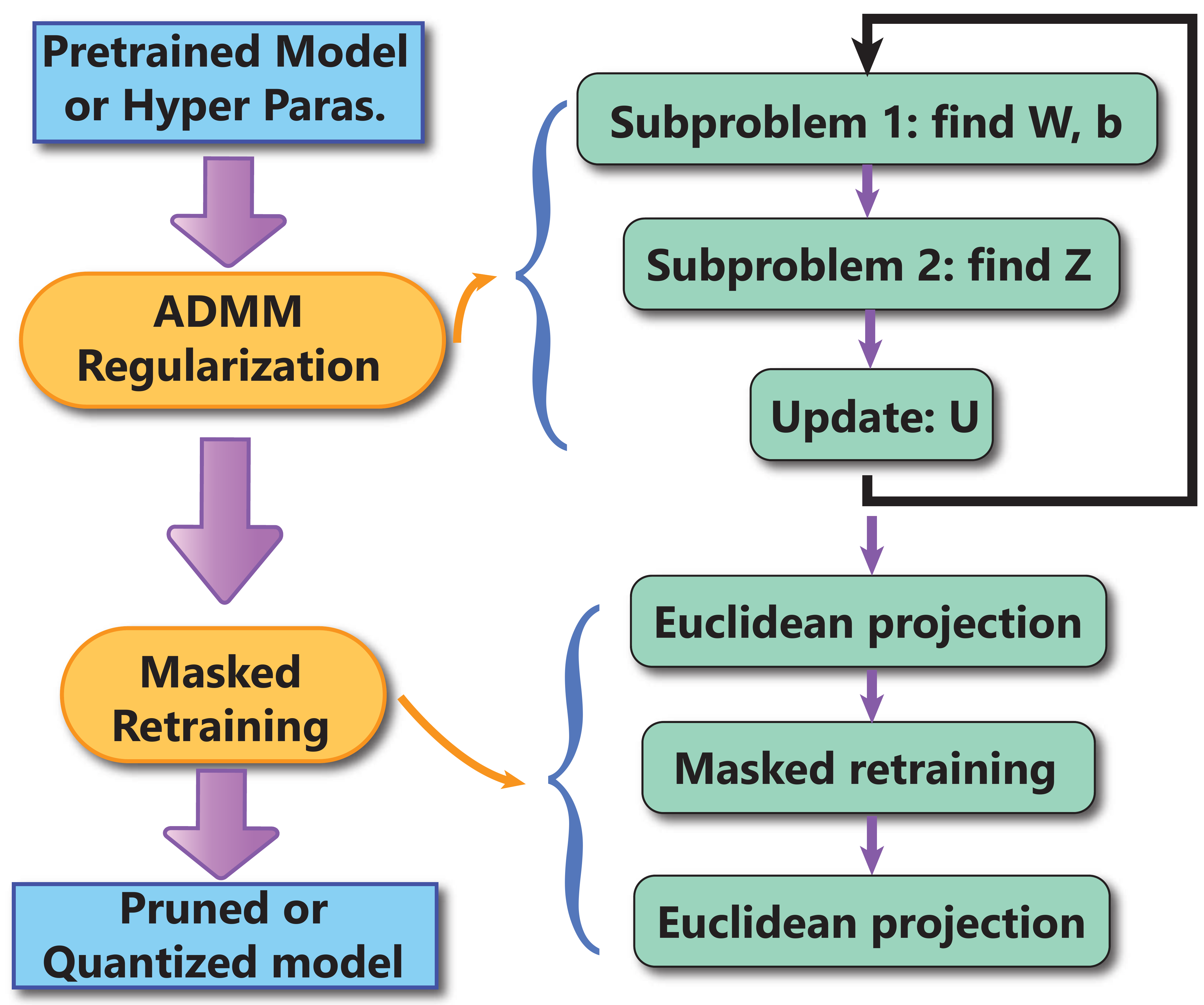}  
      \vspace{-1mm}
     \caption{Procedure of ADMM-NN-S. }
     \label{fig:admm_general}
      \vspace{-2mm}
 \end{figure}

Despite such compatibility, it is still challenging 
to directly apply ADMM due to the non-convexity in objective function. To overcome this challenge, we propose dynamic ADMM regularization, masked mapping and retraining steps for both
non-structured and structured pruning. By integrating these techniques, ADMM-NN-S can guarantee solution feasibility (satisfying all constraints) and provide high solution quality (pruning/quantization rate under the same accuracy).
The procedure of ADMM-NN-S is shown in 
Figure \ref{fig:admm_general}.

\emph{\textbf{ADMM Regularization Step}}: The ADMM regularization decomposes the original problem (\ref{opt0}) into two subproblems through\footnote{The details of ADMM are presented in \cite{boyd2011distributed,zhang2018systematic}. We omit the details due to space limitation.} (i) defining indicator function

$g_{i}({\bf{W}}_{i})=
\begin{cases}
 0 & \text { if } {\bf{W}}_{i}\in {\mathcal{S}}_{i}, \\ 
 +\infty & \text { otherwise}
\end{cases}$ 

\noindent corresponding to every set ${\mathcal{S}}_{i}$; (ii) incorporating auxiliary variables ${\bf{Z}}_{i}$, $i = 1, \ldots, N$; and (iii) adopting augmented Lagrangian \cite{boyd2011distributed}. These decomposed subproblems will be iteratively solved until convergence. The first subproblem is
\begin{equation}
\label{subproblem_1}
 \underset{ \{{\bf{W}}_{i}\},\{{\bf{b}}_{i} \}}{\text{minimize}}
\ \ \ f \big( \{{\bf{W}}_{i} \}_{i=1}^N, \{{\bf{b}}_{i} \}_{i=1}^N \big)+\sum_{i=1}^{N} \frac{\rho_{i}}{2}  \| {\bf{W}}_{i}-{\bf{Z}}_{i}^{k}+{\bf{U}}_{i}^{k} \|_{F}^{2}, \\
\end{equation}
where ${\bf{U}}_{i}^{k}:={\bf{U}}_{i}^{k-1}+{\bf{W}}_{i}^{k}-{\bf{Z}}_{i}^{k}$. The first term in the objective function of (\ref{subproblem_1}) is the differentiable loss function of the DNN, and the second term is a quadratic regularization term of the ${\bf{W}}_{i}$'s, which is differentiable and convex. As a result (\ref{subproblem_1}) can be solved by stochastic gradient descent as original DNN training. Please note that this first subproblem maintains the same form and solution for (non-structured and structured) weight pruning and quantization problems.

The second subproblem is given by
\begin{equation}
 \underset{ \{{\bf{Z}}_{i} \}}{\text{minimize}}
\ \ \ \sum_{i=1}^{N} g_{i}({\bf{Z}}_{i})+\sum_{i=1}^{N} \frac{\rho_{i}}{2} \| {\bf{W}}_{i}^{k+1}-{\bf{Z}}_{i}+{\bf{U}}_{i}^{k} \|_{F}^{2}. \\
\end{equation}

Note that $g_{i}(\cdot)$ is the indicator function of ${\mathcal{S}}_{i}$, thus this subproblem can be solved analytically and optimally \cite{boyd2011distributed}. For $i = 1, \ldots, N$, the optimal solution is the Euclidean projection of ${\bf{W}}_{i}^{k+1}+{\bf{U}}_{i}^{k}$ onto ${\mathcal{S}}_{i}$. 
For \emph{non-structured weight pruning}, we can prove that the Euclidean projection results in keeping $\alpha_i$ elements in ${\bf{W}}_{i}^{k+1}+{\bf{U}}_{i}^{k}$ with the largest magnitudes and setting the remaining weights to zeros. 
For \emph{filter pruning}, we first calculate $O_{a}=\| ({\bf{W}}_{i}^{k+1} +{\bf{U}}_{i}^{k})_{a,:,:,:} \|_F^2$ for $a = 1, \ldots, A_i$, where $\| \cdot \|_F$ denotes the Frobenius norm. We then keep $\beta_i$ elements in $({\bf{W}}_{i}^{k+1} +{\bf{U}}_{i}^{k})_{a,:,:,:}$ corresponding to the $\beta_i$ largest values in $\{O_a\}_{a = 1}^{A_i}$ and set the rest to zero.
For \emph{channel pruning}, we first calculate $O_{b}=\| ({\bf{W}}_{i}^{k+1} +{\bf{U}}_{i}^{k})_{:,b,:,:} \|_F^2$ for $b= 1, \ldots, B_i$. We then keep $\gamma_{i}$ elements in $({\bf{W}}_{i}^{k+1} +{\bf{U}}_{i}^{k})_{:,b,:,:}$ corresponding to the $\gamma_{i}$ largest values in $\{O_b\}_{b=1}^{B_i}$ and set the rest to zero.
The optimal solution of the second subproblem for \emph{filter shape pruning} is similar, and is omitted due to space limitation.
For \emph{weight quantization}, we can prove that the Euclidean projection results in mapping every element of ${\bf{W}}_{i}^{k+1}+{\bf{U}}_{i}^{k}$ to the quantization level closest to that element.

After both subproblems solved, we update the dual variables ${\bf{U}}_{i}$'s according to the ADMM rule \cite{boyd2011distributed} and thereby complete one iteration in ADMM regularization. 
Overall the ADMM regularization step can be understood as a smart, dynamic $L_2$ regularization, in which the regularization target $ {\bf{Z}}_{i}^{k}-{\bf{U}}_{i}^{k}$ will change judiciously and analytically in each iteration. On the other hand, conventional regularization methods (based on $L_1$, $L_2$ norms or their combinations) use a fixed regularization target, and the penalty is applied on all the weights. This will inevitably cause accuracy degradation. Sample comparison results are in Section \ref{sec:results1}. 

\emph{\textbf{Masked mapping and retraining}}: After ADMM regularization, we obtain intermediate ${\bf{W}}_{i}$ solutions. The subsequent step of masked mapping and retraining will guarantee the solution feasibility and improve solution quality. For non-structured and structured weight pruning, the procedure is more straightforward. We first perform the said Euclidean projection (mapping) to guarantee that pruning constraints are satisfied. Next, we mask the zero weights and retrain the DNN with non-zero weights using training sets, while keeping the masked weights 0. In this way test accuracy (solution quality) can be (partially) restored, and solution feasibility (constraints) will be maintained.

For weight quantization, the procedure is more complicated. The reason is that the retraining process will affect the quantization results, thereby solution feasibility. To deal with this issue, we first perform Euclidean projection (mapping) of weights that are close enough (defined by a threshold value $\epsilon$) to nearby quantization levels. Then we perform retraining on the remaining, unquantized weights (with quantized weights fixed) for accuracy improvement. Finally we perform Euclidean mapping on the remaining weights as well. In this way the solution feasibility will be guaranteed.

\subsection{Techniques for Enhancing Convergence}

In this section we discuss two techniques for enhancing convergence (rate and results): multi-$\rho$ method in ADMM regularization, and progressive weight pruning. We abandon the extragradient descent method in \cite{leng2017extremely} as we did not find the advantage in convergence speed, not to mention the additional hyperparameters introduced by this method.

\emph{\textbf{Increasing $\rho$ in ADMM regularization}}: The $\rho_{i}$ values are the most critical hyperparameter in ADMM regularization. We start from smaller $\rho_{i}$ values, say $\rho_{1} = \dots = \rho_{N} = 1.5\times 10^{-3}$, and gradually increase with ADMM iterations. This coincides with the theory of ADMM convergence \cite{hong2016convergence,liu2018zeroth}. It in general takes 8 - 12 ADMM iterations for convergence, corresponding to 100 - 150 epochs in PyTorch. This convergence rate is comparable with the original DNN training.

\emph{\textbf{Progressive weight pruning}}: The ADMM regularization is $L_2$ regularization. As a result, there is a large portion of very small weights values after one round of ADMM-based (non-structured or structured) weight pruning. This gives rise to the opportunity to perform a second round of weight pruning. In practice, we perform \emph{two rounds} of ADMM-based weight pruning consecutively, where the weight pruning results in the first round will be the starting point of the second round (weights that are already pruned to zero will not be recovered). This method has an additional benefit of reducing the search space in each step, thereby accelerating convergence.

\subsection{Determining Hyperparameters}\label{sec:hyper_deter}

Hyperparameter determination mainly refers to the determination process of pruning rate (e.g., the $\alpha_i$ value) and/or the number of quantization levels per layer of DNN. This is a more challenging task for pruning than quantization in general. For quantization, it is typically preferred for the same number of quantization levels for all (or most of) layers, like binarized or ternarized weights, which is preferred by hardware. For weight pruning, on the other hand, these pruning rate values are flexible and shall be judiciously determined.

As hyperparameter determination is not our primary focus, we use a heuristic method as follows. We observe that we can achieve at least 3$\times$ more weight pruning than prior, heuristic weight pruning methods without accuracy loss. Hence, we adopt the per-layer pruning rates reported in prior work, and increase proportionally. In the progressive pruning procedure, we set the target of the first round to be 1.5$\times$ pruning than prior work, and the second round to be doubled based on that. We will further increase the pruning rates if there is still margin for weight pruning without accuracy loss.

\section{Non-structured DNN Weight Pruning and Quantization Results}\label{sec:results1}

In this section, we demonstrate the effectiveness of ADMM-NN-S for non-structure pruning and quantization, based on ImageNet ILSVRC-2012, CIFAR-10, and MNIST data sets, using AlexNet \cite{krizhevsky2012imagenet}, VGGNet \cite{simonyan2014very}, ResNet-18/ResNet-50 \cite{he2016deep}, MobileNet V2 \cite{sandler2018mobilenetv2}, and LeNet-5 DNN models.
Due to space limitation, we only show the results 
of the overall DNN model (which has the most prior work for comparison), and binarized quantization of DNNs. 
 Our implementations are based on PyTorch, and the baseline accuracy results are in many cases higher than those utilized in prior work, which reflects the recent training advances. 
 For example, in the AlexNet model we utilize a baseline with Top-1 accuracy 60.0\% and Top-5 accuracy 82.2\%, both higher than prior work (57.2\% Top-1 and 80.2\% Top-5).
 We conduct a fair comparison because we focus on {\em relative accuracy} with our baseline instead of the absolute accuracy (which has outperformed prior work).

Thanks to the compatibility of ADMM-NN-S with DNN training, directly training a DNN model using the framework achieves the same result as using a pre-trained DNN model. When a pre-trained DNN model is utilized, we limit the number of epochs in both steps in the progressive framework to be 120, similar to the original DNN training in PyTorch and is much lower than the iterative pruning heuristic \cite{han2015learning}.

\subsection{Non-Structured Weight Pruning Results}

\emph{\textbf{AlexNet Results for ImageNet Dataset}}: Table \ref{table:AlexNet1} compares the overall pruning rates of the whole AlexNet model (CONV and FC layers) vs. accuracy, between the proposed framework and various prior methods. 
We can clearly observe that the proposed framework outperforms prior methods, including the prior ADMM method \cite{zhang2018systematic}. With almost no accuracy loss even based on
the high baseline accuracy, we achieve 36$\times$ overall pruning rate. We achieve a notable 63$\times$ weight reduction with 80.3\% Top-5 accuracy, just slightly below the baseline accuracy in prior work.

\begin{table}
\scriptsize
\centering
\vspace{-3mm}
\caption{Overall weight pruning rate comparisons on AlexNet model for ImageNet data set.}\label{table:AlexNet1}
\begin{tabular}{p{2.2cm}p{1.5cm}p{1.7cm}p{1.4cm}}
\hline
Method &  Top-5 accuracy & Relative accuracy loss  & Overall prun. rate  \\ \hline
Iter. prun. \cite{han2015learning} & $80.3$\%  & $-0.1$\% & 9.1$\times$ \\ \hline
NeST \cite{dai2017nest} & $80.3\%$ & $-0.1$\% & 15.7$\times$ \\ \hline
Dyn. surg. \cite{guo2016dynamic} & $80.0\%$ & $+0.2$\% & 17.7$\times$ \\ \hline
ADMM \cite{zhang2018systematic} & $80.2\%$ & $-0.0$\% & 17.7$\times$ \\ \hline
\bf{Our method} & $82.0\%$ & $+0.2$\% & 36$\times$ \\ \hline
\bf{Our method} & $80.8\%$ & $+1.4$\% & 44$\times$ \\ \hline
\bf{Our method} & $80.3\%$ & $+1.9$\% & 63$\times$ \\ \hline
\bf{Our method} & $77.8\%$ & $+4.4$\% & 96$\times$ \\ \hline
\end{tabular}
\end{table}

\begin{figure}[t]
\centering
\includegraphics[width=0.9\linewidth]{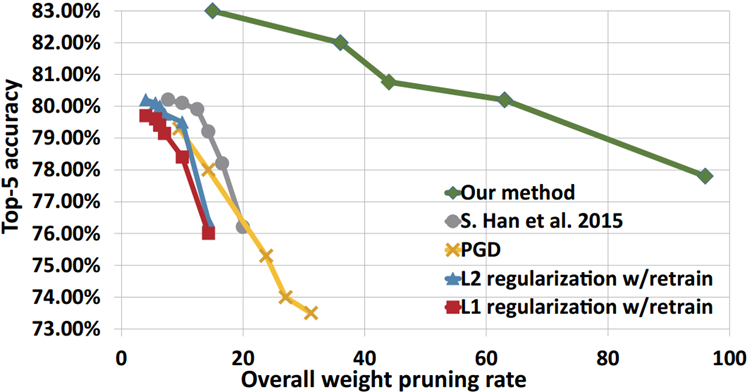}
 \vspace{-1mm}
\caption{Top-5 accuracies for different pruning methods on AlexNet for ImageNet dataset. }
\label{fig:different_alexnet}
 \vspace{-2mm}
\end{figure}

Figure \ref{fig:different_alexnet} illustrates the absolute top-5 accuracy for different pruning methods, on AlexNet model for ImageNet dataset. These methods include our proposed solution, iterative pruning \cite{han2015learning}, fixed regularization techniques like $L_1$ and $L_2$ regularizations, and projected gradient descent (PGD). The results clearly show that the proposed method outperforms the others both in absolute accuracy and in relative accuracy loss.

\emph{\textbf{ResNet-50 Results for ImageNet Dataset}}: 
Due to the lack of existing effective pruning results, we conduct uniform weight pruning, --- use the same pruning rate for all CONV and FC layers. The results are shown in Table \ref{table:ResNet-50}. We achieve 8$\times$ overall pruning rate (also 8$\times$ pruning rate on CONV layers) on ResNet-50 without accuracy loss. These results clearly outperform the prior work.

\begin{table}[ht]
\scriptsize
\centering
\vspace{-5mm}
\caption{Comparisons of overall weight pruning results on ResNet-50 for ImageNet data set.}\label{table:ResNet-50}
\begin{tabular}{p{3cm}p{2.5cm}p{1.8cm}}
\hline
Method & Top-5 Acc. Loss  & Pruning rate \\ 
\hline
Uncompressed & 0.0\%  & 1$\times$ \\ \hline
{Fine-grained \cite{mao2017exploring}}  & 0.1\%  & 2.6$\times$ \\ \hline
{ADMM-NN~\cite{ren2019ADMMNN}}  & 0.0\%  & 7$\times$ \\ \hline
\bf{Our method}  & 0.0\%  & 8$\times$ \\ \hline
\bf{Our method}  & 0.7\%  & 17.4$\times$ \\ \hline
\end{tabular}
\end{table}

\emph{\textbf{MobileNet V2 Results for CIFAR-10 Dataset}}: The baseline accuracy is as high as 95.07\% due to the adoption of mixup technique. We present our results in Table \ref{table:MobileNet} due to the lack of prior work for fair comparison. We achieve 5.7$\times$ weight pruning with almost no accuracy loss, starting from the high-accuracy baseline. We achieve 10$\times$ weight pruning (which is highly challenging for MobileNet) with only 1.3\% accuracy loss.

\begin{table}[ht]
\scriptsize
\centering
\vspace{-3mm}
\caption{Our weight pruning results on MobileNet V2 for CIFAR-10 data set.}\label{table:MobileNet}
\begin{tabular}{p{3cm}p{2.5cm}p{1.8cm}}
\hline
Method & Accuracy  & Pruning rate \\ 
\hline
Uncompressed & 95.07\%  & 1$\times$ \\ \hline
\bf{Our method}  & 94.95\%  & 5.7$\times$ \\ \hline
\bf{Our method}  & 94.70\%  & 6.7$\times$ \\ \hline
\bf{Our method}  & 93.75\%  & 10$\times$ \\ \hline
\end{tabular}
\vspace{-2mm}
\end{table}

\emph{\textbf{LeNet-5 Results for MNIST Dataset}}: Table \ref{table:LeNet-5} demonstrates the comparison results on LeNet-5 model using MNIST data set. We achieve an unprecedented 348$\times$ overall weight reduction with almost no accuracy loss. It clearly outperforms prior methods including one-shot ADMM-based method  \cite{zhang2018systematic}.

\begin{table}[ht]
\scriptsize
\centering
\vspace{-5mm}
\caption{Comparisons of overall weight pruning results on LeNet-5 for MNIST data set.}\label{table:LeNet-5}
\begin{tabular}{p{3.5cm}p{1.5cm}p{1.8cm}}
\hline
Method & Accuracy &  Pruning rate \\ 
\hline
Uncompressed & 99.2\% &  1$\times$ \\ \hline
 {Network Pruning \cite{han2015learning}}  & 99.2\% &  12.5$\times$ \\ \hline
 {ADMM \cite{zhang2018systematic}}  & 99.2\% &  71.2$\times$ \\ \hline
\bf{Our method}  & 99.2\% &  246$\times$ \\ \hline 
\bf{Our method}  & 99.0\% &  348$\times$ \\ \hline
\end{tabular}
\vspace{-6mm}
\end{table}

\subsection{Binary Weight Quantization Results}

Due to space limitation, we mainly show the results on fully binarized DNN models (i.e., weights in all layers, including the first and the last, are binarized), which is a highly challenging task. Please note that the amount of prior work on fully binarized weight quantization is very limited due to the highly challenging nature.

\emph{\textbf{Weight Quantization Results on LeNet-5 and CIFAR-10}}: To the best
of our knowledge, we achieve {\em the first lossless, fully binarized LeNet-5 model}. The accuracy is still 99.21\%, lossless compared with baseline. 
In prior works, achieving lossless is challenging even for MNIST. 
For example, recent work \cite{cheng2018differentiable} results in 2.3\% accuracy degradation on MNIST for full binarization, with baseline accuracy 98.66\%.
 We also achieve \emph{the first lossless, fully binarized VGG-16 for CIFAR-10}. The accuracy is 93.53\%. We would like to point out that fully ternarized quantization results in 93.66\% accuracy. Table \ref{table:VGG-quan} shows our results and comparisons.

\begin{table}[ht]
\centering
\scriptsize
\vspace{-5mm}
\caption{Comparisons of fully binary (ternary) weight quantization results on VGG-16 for CIFAR-10 data set.}\label{table:VGG-quan}
\begin{tabular}{p{2.5cm}p{1.5cm}p{2.5cm}}
\hline
Method & Accuracy &  Num. of bits \\ 
\hline
Baseline of \cite{cheng2018differentiable} & 84.80\% &  32 \\ \hline
 Binary \cite{cheng2018differentiable}  & 81.56\% &  1 \\ \hline
 \bf{Our baseline}  & 93.70\% &  32 \\ \hline
 \bf{Our ternary}  & 93.66\% &  2 (ternary) \\ \hline
\bf{Our binary}  & 93.53\% &  1 \\ \hline 
\end{tabular}
\vspace{-1mm}
\end{table}

\emph{\textbf{Binary Weight Quantization Results on ResNet for ImageNet}}: 
The binarization of ResNet models on ImageNet data set is widely acknowledged as an
extremely challenging task. As a result, there are very limited prior work (e.g., the prior ADMM-based method \cite{leng2017extremely}) with binarization results on ResNet models. As \cite{leng2017extremely} targets ResNet-18, we make a fair comparison on the same model. Table \ref{table:ResNet-quan} demonstrates the comparison results (Top-5 accuracy loss).
 In prior work, by default the first and last layers are not quantized (to 8 bits) as these layers have a significant effect on overall accuracy. When leaving the first and last layers unquantized, we observe the higher accuracy compared with the prior method.
The Top-1 accuracy has similar result: 3.8\% degradation in our method and 4.3\% in \cite{leng2017extremely}.

Furthermore, we can derive a \emph{fully binarized ResNet-18}, in which weights in all layers are binarized. The accuracy degradation is 5.8\%, which is noticeable and shows that the full binarization of ResNet is a challenging task even for the proposed framework. We did not find prior work to compare with this result.

\begin{table}[ht]
\centering
\scriptsize
\vspace{-5mm}
\caption{Comparisons of weight quantization results on ResNet-18 for ImageNet data set.}\label{table:ResNet-quan}
\begin{tabular}{p{2.7cm}p{2cm}p{2.2cm}}
\hline
Method & Relative Top-5 acc. loss  & Num. of bits \\ 
\hline
Uncompressed & 0.0\%  & 32 \\ \hline
 {ADMM \cite{leng2017extremely}}  & 2.9\%  & 1 (8 for the first and last) \\ 
\hline
\bf{Our method}  & 2.5\%  & 1 (8 for the first and last)  \\ \hline
\bf{Our method}  & 5.8\%  & 1 \\ \hline
\end{tabular}
\vspace{-4mm}
\end{table}

\noindent
{\bf Summary}
The results presented in this section
show that ADMM-NN-S can achieve
better results compared to
state-of-the-art. In certain cases, 
ADMM-NN-S achieves unprecedented weight reduction.
These results provide a strong baseline 
and credibility of our study.

\section{Non-structured vs. Structured: The Comparisons Methodology}

\textbf{\emph{A Motivation Example:}} 
The previous section has shown the
superior results on joint weight pruning and quantization.
Using LeNet-5 (MNIST data set) as an example, we achieve an unprecedented 348$\times$ non-structured weight reduction together with 3-bit quantization, maintaining 99$\%+$ accuracy. When indices are not accounted for, the overall compression rate is an unprecedented 3,712$\times$ compared with the original LeNet-5 model without compression. 
However, each index needs to be at least 9-bit considering 348$\times$ weight pruning. This makes index storage even larger than weights, and indices cannot be further quantized. As a result, \emph{non-structured weight pruning in fact 
results in larger actual storage than structured pruning}. 

The fundamental phenomena shown here is that,
with quantization the 
weight reduction by non-structured pruning is 
offset by the extra index storage.
It motivates us to study whether it is a common
trend with weight quantization in place?
If the answer is yes, then the value of non-structured weight pruning will be further 
in doubt. This is because non-structured pruning is already less preferred for GPU and CPUs \cite{wen2016learning,yu2017scalpel}, 
the only benefit is the potentially higher pruning rates due to 
greater pruning flexibility.
 If this benefit is also lost, there will be nearly no merit of non-structured sparsity for hardware acceleration of DNNs, considering 
 the impacts on computation efficiency and 
 degraded parallelism. 
 Importantly, such conclusion will also be true for FPGA and ASIC and guide us to the 
 design aspects that we should really focus on.
 
 In this section, we conduct the {\em first(to the best of our knowledge) comprehensive study
 to understand the value of non-structured and
 structured pruning, with quantization in place
 and the same accuracy}. 
 It is worth noting that without ADMM-NN-S
 framework, this study is {\em not possible}, 
 --- we need a framework that achieves 
 competitive results and can jointly
 perform both weight pruning and quantization.


\textbf{\emph{A Hardware Implementation-Agnostic Comparison Methodology:}} 
We conduct a fair comparison between non-structured and structured weight pruning with quantization in place, based on the unified solution framework. Note that the comparison framework is more FPGA and ASIC oriented as flexible weight quantization is assumed. However, we would like to point out that a moderate, fixed weight quantization, e.g., 8 bit, supported in GPU \cite{paszke2017pytorch}, TPU \cite{tpu}, and mobile devices \cite{TensorFlow-Lite}, will result in a similar conclusion.  

The key characteristic of our comparison framework is that it is \emph{hardware implementation-agnostic}. Our intention is that the comparison results will be independent of specific hardware implementations, and as a result, the conclusion will unlikely to change for architectural advances in either type of pruning. 
Therefore, we directly compare the amounts of \emph{storage} and estimated \emph{computation 
efficiency} for non-structured and structured weight pruning with quantization in place, which 
capture the fundamental trade-offs. Intuitively, storage is measured as the total weight and index storage with quantization in place. Storage of intermediate results is not considered, and this favors non-structured pruning, --- structured, filter/channel pruning will likely benefit more in intermediate results storage reduction.

On the other hand, computation efficiency is estimated using the \emph{pruning-to-performance ratio} (PPR) values derived from prior work on non-structured sparsity accelerators \cite{yuan2018sticker,ren2018admm,zhang2016cambricon,parashar2017scnn}. For structured pruning, $\alpha\times$ weight reduction results in around $\alpha\times$ speedup (slightly higher or lower depending on platform and problem), and the PPR value is approximately 1. For non-structured pruning, $\alpha\times$ weight reduction only results in $\beta\times$ speedup with $\beta<\alpha$. In the state-of-art tapeouts \cite{yuan2018sticker}, the PPR value $\alpha/\beta>3$, which is close to 3 with a low pruning rate and higher than 4 for a high pruning rate. In synthesis results \cite{ren2018admm,zhang2016cambricon,parashar2017scnn}, this PPR value ranges from 2.7 to 3.5. We use the smallest value 2.7 that favors non-structured pruning the most. In other words, if non-structured pruning achieves more than 2.7$\times$ pruning rate than structured one (or equivalently, structured pruning rate is less than 37\% of non-structured one) under the same accuracy and quantization level, the former is more preferred in terms of computation. Otherwise, the latter is more preferred.

\begin{figure}
    \centering
    \includegraphics[width=0.45 \textwidth]{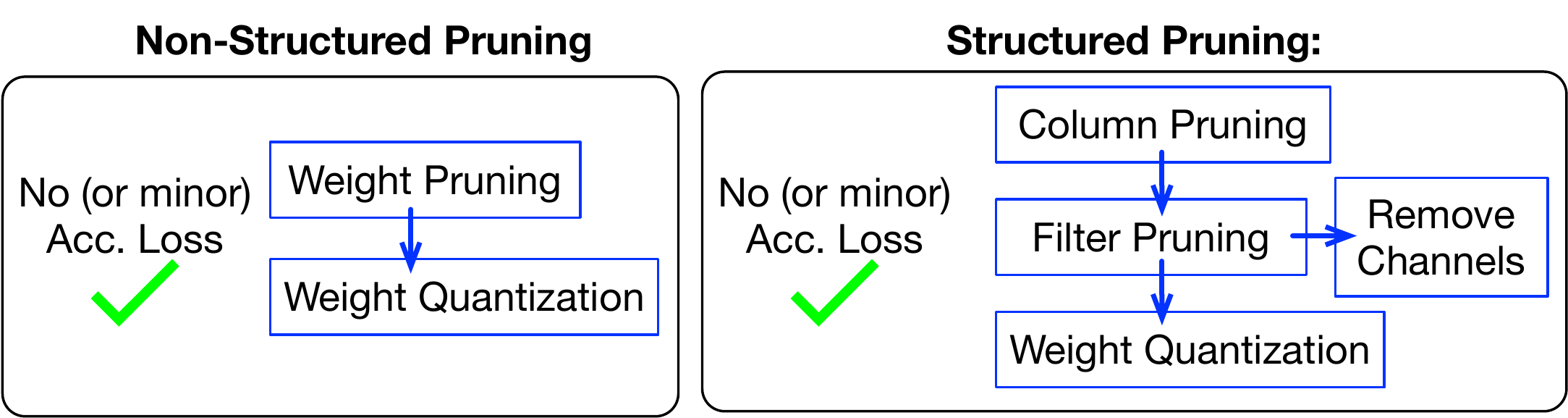}
    \vspace{-1mm}
    \caption{Procedure for maintaining accuracy.}
    \label{fig:keepaccuracy}
    \vspace{-2mm}
\end{figure}

\textbf{\emph{Maintaining the Same Accuracy for Comparison:}} 
The proposed comparison is performed under the same accuracy for non-structured and structured pruning with quantization in place. The {\em precise accuracy control}, which is challenging for prior work, is enabled by the unified solution framework. For most cases, we would like to have (almost) no accuracy degradation compared with the baseline DNN model without pruning or quantization. For non-structured pruning, it is achieved in two steps: 1) perform weight pruning to the maximum extent such that there will be no accuracy loss; and 
2) perform weight quantization (hopefully) not to cause accuracy loss. For structured pruning, we give priority to column pruning, and perform three steps: 1) perform column pruning to the maximum extent without accuracy loss; 2) perform filter pruning and reduce corresponding redundant channels; and 3) perform weight quantization (hopefully) without accuracy loss. \textcolor{black}{Figure \ref{fig:keepaccuracy}} illustrates the procedure for maintaining accuracy. Of course the proposed framework is also applicable if certain accuracy degradation is allowed. 
A larger margin of accuracy loss in general favors structured pruning,
because higher pruning rates can be achieved for both pruning schemes, 
but non-structured pruning requires more bits for indices. 



There is more subtlety in the combination of non-structured pruning and quantization. If a weight is non-zero after pruning but quantized to zero, this weight can be added to the pruned list to achieve a higher pruning rate. Please note that this phenomenon does {\em not} apply to structured pruning. To better exploit this phenomenon and achieve even higher storage/computation reduction for non-structured pruning (plus quantization), we leverage the state-of-art ternary quantization technique \cite{he2018simultaneously} with dedicated optimizations. We apply this technique for weight quantization after non-structured pruning in cases when it outperforms our proposed method, thereby providing enough opportunity to non-structured weight pruning.

\section{Comparison of Non-Structured and Structured Weight Pruning}

Due to space limitation, we focus on CONV layers, which are the most computationally intensive layers in DNNs and are becoming the major storage in state-of-art ResNet and MobileNet models. We do observe similar (and more significant) effect on FC layers and on RNNs, without providing detailed results due to space. 

As discussed in Section \ref{sec:results1}, our implementations are based on PyTorch with high baseline accuracies. We limit the number of epochs in both structured pruning and non-structured pruning to be 240 (much lower than the iterative pruning heuristic \cite{han2015learning}), and the number of epochs in weight quantization to be 120. We adopt hyperparameter determination heuristic discussed in Section \ref{sec:hyper_deter} for both structured and non-structured pruning.

For non-structured weight pruning, we show results on both CSR with relative indices and with absolute indices. The former is more appropriate for storage reduction, but the latter achieves higher
computation efficiency. For absolute indices we assume 4K$=64\times 64$ blocks that are reasonable for hardware \cite{yuan2018sticker}. Besides the comparison between two pruning schemes, our results also consistently outperform prior work, in terms of both non-structured and structured pruning, as well as combination with weight quantization.

\subsection{Comparison Results on ImageNet Dataset}

Table \ref{table:results_AlexNet} and Table \ref{table:results_resnet18_imagenet} demonstrate the comparison results using AlexNet and ResNet-18 models on ImageNet dataset.
In these tables, ``CONV Prune Rate" refers to the reduction ratio in the number of weights in overall CONV layers, and the number of remaining weights is ``CONV No. of Weights". ``CONV Quant Bits" refers to the number of bits used for equal-distance weight quantization, while ``CONV Weight Store" is the storage required only for weights (not account for indices). ``Index Bits" refers to the number of bits in CSR with relative indices. In our results, we already optimized this index bit value to minimize the overall storage (accounting for the additional dummy zeros as well). The next two columns refer to the total storage size accounting for relative indices and absolute indices, respectively. For structured pruning, they are the same as weight storage. The final column ``CONV Compress Rate" refers to the storage compression rate compared with the original baseline DNN model without compression, assuming relative indices that are more favorable to non-structured pruning. We use ``N/A" if the specific prior work only focuses on weight pruning without performing quantization.

 It can be observed that we achieve significant pruning rate gains for both non-structured and structured pruning. Especially for structured pruning, we achieve 5.1$\times$ and 2.5$\times$ structured weight pruning in CONV layers of AlexNet and ResNet-18 models, respectively, without accuracy loss. We further achieve 4.3$\times$ structured pruning with minor accuracy loss around 1\%. For ResNet on ImageNet dataset, it is difficult for prior work to achieve lossless structured pruning. For example, \cite{he2017channel} results in 1\% accuracy loss with 2$\times$ structured pruning, on ResNet-50 model with more redundancy.

When comparing non-structured vs. structured pruning, the overall CONV compression rate is comparable for the AlexNet case and the 1\% accuracy loss case for ResNet-18. For the lossless case in ResNet-18, non-structured pruning is slightly better in storage, especially when relative indices are utilized. This is because the number of bits for indexing is relatively small in this case, and the slight benefit will diminish if certain accuracy loss is tolerable. The occasional gain cannot outweigh the difficulty in hardware support of non-structured sparsity. It would be difficult to choose non-structured pruning over the other one 
even if the storage results are comparable.

\begin{table*}[!t]
    \centering
    \caption{Comparison on Non-Structured vs. Structured Pruning using AlexNet on ImageNet Dataset}
    \renewcommand{\arraystretch}{1}
    \resizebox{\textwidth}{!}{
        \begin{tabular}{c c c c c c c c c c c} 
					\hline\hline
					\multirow{2}{*}{} & \multirow{2}{*}{Method} & \multirow{2}{*}{\makecell{Top-5 \\ Accuracy}} & \multirow{2}{*}{\makecell{CONV \\ Prune Rate}} & \multirow{2}{*}{\makecell{CONV No.\\ of Weights}} & \multirow{2}{*}{\makecell{CONV \\ Quant Bits}} & \multirow{2}{*}{\makecell{CONV \\ Weight Store}} & \multirow{2}{*}{\makecell{Index \\ Bits}} & \multirow{2}{*}{\makecell{Weight+Index \\ Storage (Relative)}} & \multirow{2}{*}{\makecell{Weight+Index \\ Storage (Absolute)}} & \multirow{2}{*}{\makecell{CONV \\ Compress Rate}}\\ \\
					\hline
					\bf{Baseline AlexNet} &  & 82.2\% & 1.0$\times$ & 2.3M & 32 & 9.3MB & - & 9.3MB & 9.3MB & 1.0$\times$ \\ 
					\hline
					\multirow{5}{*}{\makecell{\textbf{Non-} \\ \textbf{structured}}} & Han~\cite{han2015deep} & 80.3\% & 2.7$\times$ & 0.86M & 8 & 0.86MB & 4 & 1.3MB & N/A & 7.1$\times$\\ 
					& Dyn. surg. \cite{guo2016dynamic} & 80.0\% & 3.1$\times$ & 0.74M & N/A & N/A & N/A & N/A & N/A &N/A \\ 
					 & Nest \cite{dai2017nest} & 80.3\% & 3.23$\times$ & 0.71M & N/A & N/A & N/A & N/A & N/A & N/A\\ 
					 & Fine-grained \cite{mao2017exploring} & 80.3\% & 4.16$\times$ & 0.55M & N/A & N/A & N/A & N/A & N/A & N/A\\ 
					 & \textbf{our's} & 81.9\% & \textbf{11.2$\times$} & 0.3M & 7 & 0.26MB & 6 & 0.51MB  & 0.61MB & \textbf{25.5}$\times$ \\ 
					 \hline
					\multirow{4}{*}{\makecell{\textbf{Structured}}} & SSL~\cite{wen2016learning} & 80.4\% & 1.4$\times$ &1.6M  & N/A & N/A & - & N/A & N/A & N/A \\ 
					& Taylor~\cite{molchanov2016pruning} & 79.8\% & 2.5$\times$ &0.92M  & N/A & N/A & - & N/A & N/A & N/A \\ 
					& NISP~\cite{yu2018nisp} & 80.2\% & 1.9$\times$ &1.2M  & N/A & N/A & - & N/A & N/A & N/A \\ 
					 & \textbf{our's} & 81.8\% & \textbf{5.1$\times$} & 0.65M & 7 & 0.56MB & - & 0.56MB & 0.56MB & \textbf{23.3$\times$} \\ 
					 \hline
					 
        \end{tabular}
            }
    \vspace{-1mm}
    \label{table:results_AlexNet}
\end{table*}

\begin{table*}[!t]
    \centering
    \caption{Comparison on Non-Structured vs. Structured Pruning using ResNet-18 (ResNet-50 in prior work NISP and ThiNet, with starting Top-5 accuracy 91.1\%), ImageNet Dataset}
    \renewcommand{\arraystretch}{1}
    \resizebox{\textwidth}{!}{
        \begin{tabular}{c c c c c c c c c c c} 
					\hline\hline
					\multirow{2}{*}{} & \multirow{2}{*}{Method} & \multirow{2}{*}{\makecell{ Accuracy}} & \multirow{2}{*}{\makecell{CONV \\ Prune Rate}} & \multirow{2}{*}{\makecell{CONV No.\\ of Weights}} & \multirow{2}{*}{\makecell{CONV \\ Quant Bits}} & \multirow{2}{*}{\makecell{CONV \\ Weight Store}} & \multirow{2}{*}{\makecell{Index \\ Bits}} & \multirow{2}{*}{\makecell{Weight+Index \\ Storage (Relative)}} & \multirow{2}{*}{\makecell{Weight+Index \\ Storage (Absolute)}} & \multirow{2}{*}{\makecell{CONV \\ Compress Rate}}\\ \\
					\hline
					\bf{Baseline ResNet-18} &  & 89.1\% & 1.0$\times$  & 11.2M  & 32 & 44.7MB & - & 44.7MB & 44.7MB & 1.0$\times$\\ 
					\hline
				    \multirow{2}{*}{\makecell{\textbf{Non-} \\ \textbf{structured}}} & \textbf{our's} & 89.1\% & \textbf{6.4}$\times$ & 1.75M & 6 & 1.32MB & 5  & 2.47MB  &3.11MB & \textbf{18.1}$\times$ \\
					& \textbf{our's} & 87.9\% & \textbf{8.9}$\times$ & 1.26M & 6 & 0.94MB & 5 & 1.89MB  &2.29MB  & \textbf{23.6}$\times$ \\ 
					 \hline
					 \multirow{7}{*}{\bf{Structured}} & DCP~\cite{zhuang2018discrimination} & 87.6\% & 2$\times$ & 5.7M & N/A & N/A & -  &  N/A & N/A & N/A \\
					& DCP~\cite{zhuang2018discrimination} & 85.7\% & 3.3$\times$ & 3.5M & N/A & N/A & -  &  N/A & N/A & N/A \\
					& ThiNet-50 \cite{luo2017thinet} & 90.7\% & 2$\times$ & 12.8M & N/A & N/A & -  & N/A  & N/A & N/A \\
					& ThiNet-30 \cite{luo2017thinet} & 88.3\% & 3.3$\times$ & 7.7M & N/A & N/A & -  & N/A  & N/A & N/A \\
					& NISP \cite{yu2018nisp} & 90.2\% & 1.8$\times$ & 14.2M & N/A & N/A & -  & N/A  & N/A & N/A \\
				    & \textbf{our's} & 89.1\% & \textbf{2.5}$\times$ & 4.46M & 6 & 3.34MB & -  & 3.34MB  & 3.34MB & \textbf{13.4}$\times$ \\
				    & \textbf{our's} & 87.8\% & \textbf{4.3}$\times$ & 2.60M & 6 & 1.95MB & - & 1.95MB & 1.95MB  & \textbf{22.9}$\times$ \\  
					 \hline
					 
        \end{tabular}
    }
    \vspace{-1mm}
    \label{table:results_resnet18_imagenet}
\end{table*}

\subsection{Comparison Results on CIFAR-10 Dataset}

Table \ref{table:results_vgg_cifar} and Table \ref{table:results_resnet18_cifar} demonstrate the comparison results using VGG-16 and ResNet-18 models on CIFAR-10 dataset. 
We observe that very significant pruning rates can be achieved compared with prior work (over 35$\times$ improvement in certain case). We investigated deeper and found that the underlying reason is the CIFAR-10 dataset itself, in that it is both ``simple'' and ``difficult''. ``Simple'' means that the input image scale is small and the number of classes is only 10; while ``difficult'' means that input images are blurred and feature extraction is not straightforward. As a result, researchers tend to migrate large-scale DNN models originally designed for ImageNet, such as VGG-16 and ResNet-18 (prior work even used ResNet-50). Consequently, there is significant margin of model compression, which can be exploited in the proposed systematic framework but difficult for heuristic methods.

Another observation is that non-structured pruning has only marginal gain in pruning rates (reduction in the number of weights) compared with structured one. Our hypothesis is that it is due to the high search space in non-structured pruning. Together with the large number of index bits due to high pruning rates, non-structured pruning is not preferable compared with structured one considering total storage size. The storage size gap is becoming surprisingly large when absolute indices are utilized.

Table \ref{table:results_mobilenet_cifar} demonstrates the comparison results using MobileNet V2 model on CIFAR-10 dataset. MobileNet is already compact and relatively difficult for further weight pruning, but we still achieve 5$\times$ structured pruning along with 4-bit quantization. Again non-structured pruning only shows minor gain in weight reduction, and it is not preferable considering indexing overheads.

\begin{table*}[!t]
    \centering
    \caption{Comparison on Non-Structured vs. Structured Pruning using VGG-16 on CIFAR-10 Dataset}
    \renewcommand{\arraystretch}{1}
    \resizebox{\textwidth}{!}{
        \begin{tabular}{c c c c c c c c c c c} 
					\hline\hline
					\multirow{2}{*}{} & \multirow{2}{*}{Method} & \multirow{2}{*}{\makecell{ Accuracy}} & \multirow{2}{*}{\makecell{CONV \\ Prune Rate}} & \multirow{2}{*}{\makecell{CONV No.\\ of Weights}} & \multirow{2}{*}{\makecell{CONV \\ Quant Bits}} & \multirow{2}{*}{\makecell{CONV \\ Weight Store}} & \multirow{2}{*}{\makecell{Index \\ Bits}} & \multirow{2}{*}{\makecell{Weight+Index \\ Storage (Relative)}} & \multirow{2}{*}{\makecell{Weight+Index \\ Storage (Absolute)}} & \multirow{2}{*}{\makecell{CONV \\ Compress Rate}}\\ \\
					\hline
					\bf{Baseline VGG-16} &  & 93.7\% & 1.0$\times$  & 14.7M  & 32 & 58.8MB & - & 58.8MB & 58.8MB & 1.0$\times$\\ 
					\hline
					\multirow{3}{*}{\bf{Non-Structured}} & Iter. prun.~\cite{han2015learning,liu2018rethinking} & 92.2\% & 2$\times$ &  $\approx$7.4M & N/A & N/A & -  & N/A & N/A & N/A \\
					& One-shot prun.~\cite{liu2018rethinking} & 92.4\% & 2.5$\times$ &  $\approx$5.9M & N/A & N/A & -  & N/A & N/A & N/A \\
					& \textbf{our's} & 93.1\% & \bf{57.4}$\times$ &  0.26M & 5 & 0.16MB & 7  & 0.54MB & 0.72MB & \bf{109}$\times$ \\
					 \hline
					\multirow{4}{*}{\makecell{\textbf{Structured}}} & 2PFPCE~\cite{min20182pfpce} & 92.8\% & $4\times$ & 3.7M & N/A &N/A  & - & N/A  & N/A & N/A  \\ 
					& 2PFPCE~\cite{min20182pfpce} & 91.0\% & 8.3$\times$ & 1.8M & N/A & N/A &-  &N/A  & N/A & N/A \\ 
					& ConvNet~\cite{li2016pruning} & 93.4\% & 2.7$\times$ &  5.3M & N/A & N/A & -  & N/A & N/A & N/A \\
					 & \textbf{our's} & 93.1\% & \textbf{50.0}$\times$ & 0.29M & 5 & 0.18MB &-  &0.18MB  & 0.18MB & \textbf{327}$\times$ \\ 
					 \hline
					 
        \end{tabular}
    }
    \vspace{-1mm}
    \label{table:results_vgg_cifar}
\end{table*}

\begin{table*}[!t]
    \centering
    \caption{Comparison Results on Non-Structured vs. Structured Pruning using ResNet-18 (ResNet-50 in prior work AMC and ResNet-56 in prior work NISP) on CIFAR-10 Dataset}
    \renewcommand{\arraystretch}{1}
    \resizebox{\textwidth}{!}{
        \begin{tabular}{c c c c c c c c c c c} 
					\hline\hline
					\multirow{2}{*}{} & \multirow{2}{*}{Method} & \multirow{2}{*}{\makecell{ Accuracy}} & \multirow{2}{*}{\makecell{CONV \\ Prune Rate}} & \multirow{2}{*}{\makecell{CONV No.\\ of Weights}} & \multirow{2}{*}{\makecell{CONV \\ Quant Bits}} & \multirow{2}{*}{\makecell{CONV \\ Weight Store}} & \multirow{2}{*}{\makecell{Index \\ Bits}} & \multirow{2}{*}{\makecell{Weight+Index \\ Storage (Relative)}} & \multirow{2}{*}{\makecell{Weight+Index \\ Storage (Absolute)}} & \multirow{2}{*}{\makecell{CONV \\ Compress Rate}}\\ \\
					\hline
					\bf{Baseline ResNet-18} &  & 93.9\% & 1.0$\times$  & 11.2M  & 32 & 44.6MB & - & 44.6MB & 44.6MB & 1.0$\times$\\ 
					\hline
					\bf{Non-Structured} & \textbf{our's} & 93.3\% & \bf{69.0}$\times$ & 0.16M  & 5 & 0.10MB & 8 & 0.33MB & 0.53MB & \bf{135}$\times$ \\ 
					 \hline
					\multirow{3}{*}{\makecell{\textbf{Structured}}} & AMC~\cite{He_2018_ECCV} & 93.5\% & 1.7$\times$ & N/A & N/A &N/A  & - & N/A  & N/A & N/A  \\ 
					& NISP~\cite{yu2018nisp} & 93.2\% & 1.7$\times$ & N/A & N/A &N/A  & - & N/A  & N/A & N/A  \\
					 & \textbf{our's} & 93.3\% & \textbf{59.8$\times$} & 0.19M & 5 & 0.12MB &-  &0.12MB  & 0.12MB & \textbf{372$\times$} \\ 
					 \hline
					 
        \end{tabular}
    }
    \vspace{-1mm}
    \label{table:results_resnet18_cifar}
\end{table*}

\begin{table*}[!t]
    \centering
    \caption{Comparison Results on Non-Structured vs. Structured Pruning using MobileNet-V2 on CIFAR-10 Dataset}
    \renewcommand{\arraystretch}{1}
    \resizebox{\textwidth}{!}{
        \begin{tabular}{c c c c c c c c c c c} 
					\hline\hline
					\multirow{2}{*}{} & \multirow{2}{*}{Method} & \multirow{2}{*}{\makecell{ Accuracy}} & \multirow{2}{*}{\makecell{CONV \\ Prune Rate}} & \multirow{2}{*}{\makecell{CONV No.\\ of Weights}} & \multirow{2}{*}{\makecell{CONV \\ Quant Bits}} & \multirow{2}{*}{\makecell{CONV \\ Weight Store}} & \multirow{2}{*}{\makecell{Index \\ Bits}} & \multirow{2}{*}{\makecell{Weight+Index \\ Storage (Relative)}} & \multirow{2}{*}{\makecell{Weight+Index \\ Storage (Absolute)}} & \multirow{2}{*}{\makecell{CONV \\ Compress Rate}}\\ \\
					\hline
					\bf{Baseline MobileNet-V2} &  & 95.1\% & 1.0$\times$  & 2.2M  & 32 & 9.0MB & - & 9.0MB & 9.0MB & 1.0$\times$\\ 
					\hline
					\bf{Non-Structured} & \textbf{our's} & 94.9\% & \bf{6.1}$\times$ & 0.37M  & 4 & 0.19MB & 4  & 0.48MB & 0.55MB & {\bf{18.8}}$\times$ \\ 
					 \hline
					\multirow{2}{*}{\bf{Structured}} & DCP~\cite{zhuang2018discrimination} & 94.7\% & 1.3$\times$ & 1.68M & N/A & N/A & -  &  N/A & N/A & N/A \\
					& \textbf{our's} & 95.1\% & \textbf{4.9$\times$} & 0.45M & 4 & 0.23MB &-  &0.23MB  & 0.23MB & \textbf{39.2$\times$} \\ 
					 \hline
					 
        \end{tabular}
    }
    \vspace{-1mm}
    \label{table:results_mobilenet_cifar}
\end{table*}

\subsection{Comparison Results on MNIST Dataset}

Table \ref{table:results_lenet_mnist} demonstrates the comparison results using LeNet-5 model on MNIST data set. It is a simple dataset, and we achieve 87.9$\times$ structured pruning on CONV layers, together with 3-bit quantization. Non-structured pruning is again not preferred due to the high index bit and marginal increase in weight reduction rate. Ironically, it results in multiple times the amount of storage compared with structured pruning, when weight quantization is in place.

\begin{table*}[!t]
    \centering
    \caption{Comparison on Non-Structured vs. Structured Pruning using LeNet-5 on MNIST Dataset}
    \renewcommand{\arraystretch}{1}
    \resizebox{\textwidth}{!}{
        \begin{tabular}{c c c c c c c c c c c} 
					\hline\hline
					\multirow{2}{*}{} & \multirow{2}{*}{Method} & \multirow{2}{*}{\makecell{ Accuracy}} & \multirow{2}{*}{\makecell{CONV \\ Prune Rate}} & \multirow{2}{*}{\makecell{CONV No.\\ of Weights}} & \multirow{2}{*}{\makecell{CONV \\ Quant Bits}} & \multirow{2}{*}{\makecell{CONV \\ Weight Store}} & \multirow{2}{*}{\makecell{Index \\ Bits}} & \multirow{2}{*}{\makecell{Weight+Index \\ Storage (Relative)}} & \multirow{2}{*}{\makecell{Weight+Index \\ Storage (Absolute)}} & \multirow{2}{*}{\makecell{CONV \\ Compress Rate}}\\ \\
					\hline
					\bf{Baseline LeNet-5} &  & 99.2\% & 1.0$\times$  & 25.5K  & 32 & 102KB & - & 102KB&102KB & 1.0$\times$\\ 
					\hline
					\multirow{2}{*}{\makecell{\textbf{Non-} \\ \textbf{structured}}} & Han~\cite{han2015deep} & 99.2\% & 7.7$\times$ &  3.33K& 8 & 3.33KB  & 5 & 7.0KB & N/A & 14.5$\times$\\ 
					 & \textbf{our's} & 99.0\% & \bf{114.3}$\times$ &  223 & 3 & 0.08KB  &  8 & 0.39KB & 0.93KB & \bf{262}$\times$ \\ 
					 \hline
					\multirow{2}{*}{\makecell{\textbf{Structured}}} & SSL~\cite{wen2016learning} & 99.0\% & 26.1$\times$ & 975 & N/A &N/A  & - & N/A  & N/A & N/A  \\ 
					 & \textbf{our's} & 99.0\% & \textbf{87.9}$\times$ & 290 & 3 & 0.11KB &-  &0.11KB  & 0.11KB & \textbf{944$\times$} \\ 
					 \hline
					 
        \end{tabular}
    }
    \vspace{-1mm}
    \label{table:results_lenet_mnist}
\end{table*}

\subsection{Comparison on Computation Efficiency}

We have shown that non-structured pruning is not preferable in terms of storage even assuming the storage-friendly CSR format with relative indices, not to mention absolute indices.
Based on our methodology, we find that 
computation efficiency shows the similar trend.


As discussed before, structured pruning will 
have higher computation efficiency if it achieves more than 37\% in the pruning rate as non-structured pruning. In all our testing, the ratio between weight pruning rates of structured vs. non-structured pruning ranges from 40\% to 87\%, with a large variation but consistently higher than 37\%. Even for the 40\% case, the choice is clear considering the difficulty in hardware design for non-structured sparsity.
As a result, we draw the final conclusion that non-structured weight pruning is in general not preferred compared with structured pruning across different platforms, application scenarios, DNN types, etc.

\section{Discussions}\label{sec:furtherdiscussion}

In this section, we discuss additional factors and variations in different platforms, and explain why our conclusion is unlikely to change. As a result, we draw the final conclusion that non-structured weight pruning is in general not preferred compared with structured pruning across different platforms, application scenarios, DNN types, etc. 

\subsection{Algorithm Improvement and Generalization Enhancement}

We consider the following question:
will our conclusion change if there is further algorithm improvement (that outperforms the ADMM-based unified solution in this paper)? 
Also, how about using a number of other recently proposed generalization enhancement techniques, such as warmup, mixup, cosine decay in \emph{bag of tricks} \cite{xie2018bag}? 
Mixup is already utilized in MobileNet V2 training in this work and can notably enhance convergence and stability in training (the original MobileNet training is very difficult). 
We hypothesize that the conclusion is likely to maintain unchanged, as these techniques are likely to enhance the results for {\em both} non-structured and structured weight pruning schemes. As the pruning rates increase, the number of bits for index representation will also increase. The results will likely even favor structured pruning to a greater extent.

\subsection{Transfer Learning and Adversarial Robustness}

In many critical applications of deep learning, such as autonomous driving and medical imaging, there is lack of sufficient labelled training data as standard image classification tasks. As a result, the \emph{transfer learning} technique \cite{pan2010survey,yosinski2014transferable,weiss2016survey} is widely applied via 
(i) pre-training a DNN model using standard data set (say ImageNet);
(ii) transferring to the target application domain; and 
(iii) performing fine tuning using target domain data. It is recently shown \cite{allen2018learning} that sufficient number of weight parameters is needed in order to maintain the generality, i.e., the ability in domain transfer. This coincides with practice that VGGNet and deep ResNets are the major types for transfer learning instead of MobileNet. From the DNN security aspects, recent work \cite{ye2019second} shows that sufficient number of parameters is required to maintain the robustness of DNN against adversarial attacks. 

We hypothesize that structured pruning may be preferred in this way because of the larger number of remaining weight parameters (compared with non-structured), which will lead
to higher probability to satisfy the generality and adversarial robustness requirements. We believe that it will be a challenge to quantify such requirements, and derive the best combination of structured pruning and quantization for performance optimization while satisfying such requirements.

\subsection{FC Layers and RNNs}

The comparison results conducted in this paper focus on CONV layers, which is the major computation part in DNNs. On the other hand, the FC layers are not negligible in DNNs. Besides, FC layers constitute major computations in \emph{recurrent neural networks} (RNNs), which is as important as convolutional neural networks \cite{tpu}. Our preliminary investigation shows that the gain of structured pruning in FC layers and in RNNs is even higher. This is an intuitive result because FC layers have higher degree of redundancy, and more number of bits for indices if non-structured pruning is utilized. It is also worth mentioning that a number of structured matrix-based techniques, such as block-circulant matrices \cite{ding2017c} and cyclic matrices \cite{deng2018permdnn}, serve as good candidates of structured pruning in FC layers. Superior results are already demonstrated in FC layers using these methods.

\subsection{Effects of Weight Quantization}\label{sec:QuantizationorNot}

In the current industry's practice, weight quantization is the major method in DNN model compression and is typically prioritized over weight pruning. As a result, it is unlikely that weight pruning is conducted alone (especially for FPGA/ASIC systems) without quantization. However, for such systems, it is possible that a fixed quantization level (or a set of levels) is utilized to accommodate different DNN models and applications, e.g., TPU supports 8 bit and 16 bit computation. Such moderate, fixed weight quantization (e.g., 8 bits) will unlikely change the general conclusion in this paper, especially accounting for the difficulty in developing dedicated hardware supporting non-structured sparsity. For GPUs, multi-core CPUs, and even mobile devices, 8-bit/16-bit weight quantization is already well supported. Structured pruning is known to be more suitable for such systems.

To the other extreme case, researchers are investigating weight quantization-only solution, including binary and ternary quantizations. As pointed out in Section \ref{sec:results1}, binary/ternary quantization can be almost lossless in many cases. However, we observe that there is still a large margin of structured pruning as shown in the compression results on CIFAR-10, and such compression rate cannot be achieved by weight quantization alone. As a result, we recommend to perform structured pruning in combination with weight quantization,



\section{Conclusion}
Non-structured and structured weight pruning
and weight quantization are major 
methods for model compression, but 
the interaction among different techniques
are never clearly understood.
This paper is the first to investigate
the value of non-structured and structured DNN weight pruning, 
when the weight quantization is in place.
We build ADMM-NN-S, a joint weight pruning and 
quantization framework with algorithmic 
supports for structured pruning, 
dynamic ADMM regulation, and 
masked mappling and retraining.
To perform fair and fundamental comparison between
non-structured and structured pruning in an implementation-agnostic manner, 
we propose a methodology that captures 
storage overhead and computation efficiency.
We perform extensive and representative
testing of ADMM-NN-S with AlexNet, VGGNet,
ResNet-18/50, MobileNet, and LeNet-5 models
based on ImageNet, CIFAR-10, and 
MNIST data sets. 
We show that ADMM-NN-S can significant
outperform the state-of-the-art results for 
non-structured pruning with quantization.
More importantly, for the first time we show that
with quantization in place and the same accuracy,
non-structured pruning is not preferable in 
terms of both storage overhead and 
computation efficiency. 
Thus, we recommend the 
community not to continue investigating DNN inference engines based on non-structured sparsity.

\section*{Acknowledgment}

This work was
supported in part by the NSF awards CNS-1739748, CCF-1937500, CCF-1919117, CCF-1901378, CCF-1919289.

\ifCLASSOPTIONcaptionsoff
  \newpage
\fi

\end{document}